%% file: paper.tex
\documentclass[english,conference, balance]{IEEEtran}
\usepackage[T1]{fontenc}
\usepackage[latin9]{inputenc}
\usepackage{array}
\usepackage{verbatim}
\usepackage{booktabs}
\usepackage{url}
\usepackage{multirow}
\usepackage{amstext}
\usepackage{graphicx}

\makeatletter

\newcommand{\noun}[1]{\textsc{#1}}
\providecommand{\tabularnewline}{\\}

\usepackage{siunitx}
\usepackage{algpseudocode}
\usepackage{layouts}
\usepackage[table]{xcolor}
\definecolor{lightgray}{gray}{0.92}

\usepackage{tikz}
\usetikzlibrary{arrows,shapes,matrix}

\@ifundefined{showcaptionsetup}{}{%
 \PassOptionsToPackage{caption=false}{subfig}}
\usepackage{subfig}
\makeatother

\usepackage{babel}
\usepackage{listings}

\begin{document}

\title{Systematic N-tuple Networks for Position Evaluation: Exceeding 90\%
in the Othello League}

\author{\IEEEauthorblockN{Wojciech Ja\'{s}kowski}\IEEEauthorblockA{Institute
of Computing Science, Poznan University of Technology, Pozna\'{n},
Poland\\
wjaskowski@cs.put.poznan.pl}}
\maketitle
\begin{abstract}
N-tuple networks have been successfully used as position evaluation
functions for board games such as Othello or~Connect Four. The effectiveness
of such networks depends on their architecture, which is determined
by the placement of constituent n-tuples, sequences of board locations,
providing input to the network. The most popular method of placing
n-tuples consists in randomly generating a small number of long, snake-shaped
board location sequences. In comparison, we show that learning n-tuple
networks is significantly more effective if they involve a~large
number of systematically placed, short, straight n-tuples. Moreover,
we demonstrate that in order to obtain the best performance and the
steepest learning curve for Othello it is enough to use n-tuples of
size just $\mathbf{2}$, yielding a network consisting of only $\mathbf{288}$
weights. The best such network evolved in this study has been evaluated
in the online Othello League, obtaining the performance of nearly
$\mathbf{96\%}$ --- more than any other player to date.

\end{abstract}

\section{Introduction\label{sec:Introduction}}

Board games have always attracted attention in AI due to they clear
rules, mathematical elegance and simplicity. Since the early works
of Claude Shannon on Chess \cite{Shannon1950Chess} and Arthur Samuel
on Checkers \cite{samuel59some}, a lot of research have been conducted
in the area of board games towards finding either perfect players
(Connect-4, \cite{allis1988knowledge}), or stronger than human players
(Othello, \cite{Buro2000OthelloMultiProbCut}). The bottom line is
that board games still constitute valuable test-beds for improving
general artificial and computational intelligence game playing methods
such as reinforcement learning, Monte Carlo tree search, branch and
bound, and (co)evolutionary algorithms.

Most of these techniques employ a position evaluation function to
quantify the value of a given game state. In the context of Othello,
one of the most successful position evaluation functions is \emph{tabular
value function} \cite{Buro1997EvalOthello} or \emph{n-tuple network}
\cite{Lucas2008Learning}. It consists of a number of $n$-tuples,
each associated with a look up table, which maps contents of $n$
board fields into a real value. The effectiveness of n-tuple network
highly depends on the placement of n-tuples \cite{Szubert2013Scalability}.
Typically, n-tuples architectures consist of a small number of long,
randomly generated, snake-shaped n-tuples \cite{Manning2012NashOthello,Szubert2013Scalability,Runarsson2014Preference}.

Despite the importance of network architecture, to the best of our
knowledge no study exist that studies and evaluates different ways
of placing n-tuples on the board.

In this paper, we propose an n-tuple network architecture consisting
of a large number of short, straight n-tuples, generated in a systematic
way. In the extensive computational experiments, we show that for
learning position evaluation for Othello, such an architecture is
significantly more effective than the one involving randomly generated
n-tuples. We also investigate how the length of n-tuples affects the
learning results. Finally, the performance of the best evolved n-tuple
network is evaluated in the online Othello League.

\section{Methods}

\subsection{Othello}

\begin{figure}
\centering{}\includegraphics{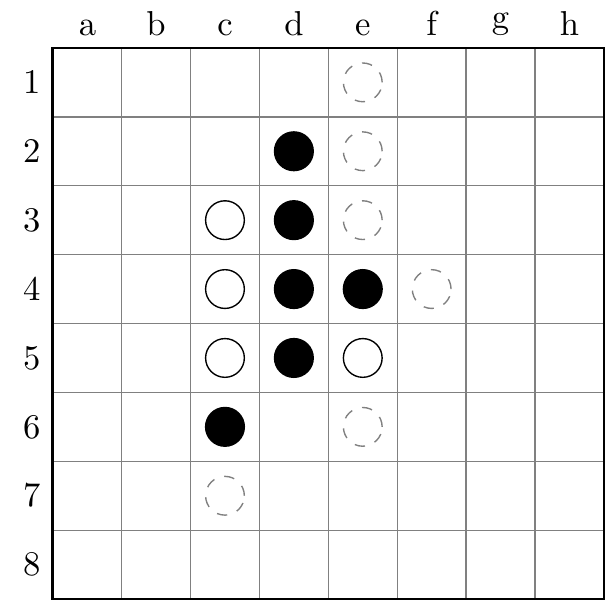} \protect\caption{\label{fig:OthelloBoard}An Othello position, where white has $6$
legal moves (dashed gray circles). If white places a piece on e$3$,
the pieces on d$3$, d$4$, and e$4$ $ $are reversed to white. }
\end{figure}

Othello (a.k.a. Reversi) is a two player, deterministic, perfect information
strategic game played on an $8\times8$ board. There are $64$ pieces
being black on one side and white on the other. The game starts with
two white and two black pieces forming an askew cross in the center
on the board. The players take turns putting one piece on the board
with their color facing up. A legal move consists in placing a piece
on a field so that it forms a vertical, horizontal, or diagonal line
with another player\textquoteright s piece, with a continuous, non-empty
sequence of opponent\textquoteright s pieces in between (see Fig.
\ref{fig:OthelloBoard}), which are reversed after the piece is placed.
Player passes if and only if it cannot make a legal move. The game
ends when both players passed consecutively. Then, the player having
more pieces with their color facing up wins.

Othello has been found to have around $10^{28}$ legal positions \cite{allis1994searching}
and has is not been solved; this is one reason why it has become such
a popular domain for computational intelligence methods \cite{lucas2007learning,Osaki2008Othello,manning2010using,Chong2012improving,Dries2012Neural,Samothrakis2012Coevolving,Szubert2013Scalability,Jaskowski2014MultiCriteria}.

\subsection{Position Evaluation Functions}

In this paper, our goal is not to design state-of-the-art Othello
players, but to evaluate position evaluation functions. That is why
our players are simple state evaluators in a $1$-ply setup: given
the current state of the board, a player generates all legal moves
and applies the position evaluation function to the resulting states.
The state gauged as the most desirable determines the move to be played.
Ties are resolved at random.

The simplest position evaluation function is position-weighted piece
counter (WPC), which is a~linear weighted board function. It assigns
a~weight $w_{rc}$ to a~board location $(r,c)$ and uses scalar
product to calculate the utility $f$ of a~board state $\mathbf{b}=(b_{rc})_{r,c=1\dots8}$:
\[
f\left(\mathbf{b}\right)=\sum_{r=1}^{8}\sum_{c=1}^{8}w_{rc}b_{rc},
\]
where $b_{ij}$ is $0$ in the case of an empty location, $+1$ if
a~black piece is present or $-1$ in the case of a~white piece.

A WPC player often used in Othello research as an expert opponent
\cite{lucas06temporal,Lucas2008Learning,szubert09coevolutionary,manning2010using,Samothrakis2012Coevolving,Szubert2013Scalability}
is Standard WPC Heuristic Player (\noun{swh}). Its weights, hand-crafted
by Yoshioka et al. \cite{yoshioka1999strategy}, are presented in
Table \ref{tab:heuristic_wpc}.

\begin{table}
\protect\caption{The weights of the Standard WPC Heuristic player (\noun{swh})\label{tab:heuristic_wpc}}

\sisetup{table-number-alignment = center, table-format=1.2}

\centering{}%
\begin{tabular}{SSSSSSSS}
1  & -.25  & .1  & .05  & .05  & .1  & -.25  & 1 \tabularnewline
-0.25  & -.25  & .01  & .01  & .01  & .01  & -.25  & -.25 \tabularnewline
.1  & .01  & .05  & .02  & .02  & .05  & .01  & .1 \tabularnewline
.05  & .01  & .02  & .01  & .01  & .02  & .01  & .05 \tabularnewline
.05  & .01  & .02  & .01  & .01  & .02  & .01  & .05 \tabularnewline
.1  & .01  & .05  & .02  & .02  & .05  & .01  & .1 \tabularnewline
-.25  & -.25  & .01  & .01  & .01  & .01  & -.25  & -.25 \tabularnewline
1  & -.25  & .1  & .05  & .05  & .1  & -.25  & 1 \tabularnewline
\end{tabular}
\end{table}

\subsection{Othello Position Evaluation Function League \label{sub:Othello-League}}

WPC is only one of the possible position evaluation functions. Others
popular ones include neural networks and n-tuple networks. To allow
direct comparison between various position evaluation functions and
algorithms capable of learning their parameters, Lucas and Runarsson
\cite{lucas06temporal} have appointed the Othello Position Evaluation
Function League %
\footnote{\url{http://algoval.essex.ac.uk:8080/othello/League.jsp}%
}. Othello League, for short, is an on-line ranking of Othello 1-ply
state evaluator players. The players submitted to the league are evaluated
against SWH (the Standard WPC Heuristic Player).

Both the game itself and the players are deterministic (with an exception
of the rare situation when at least two positions have the same evaluation
value). Therefore, to provide more continuous performance measure,
Othello League introduces some randomization to Othello. Both players
are forced to make random moves with the probability of $\epsilon=0.1$.
As a consequence the players no longer play (deterministic) Othello,
but stochastic $\epsilon$-Othello. However, it was argued that the
ability to play $\epsilon$-Othello is highly correlated with the
ability to play Othello \cite{lucas06temporal}.

The performance in Othello League is determined by the number of wins
against SWH player in $\epsilon$-Othello in $50$ \emph{double games},
each consisting of two single games played once white and once black.
To aggregate the performance into a scalar value, we assume that a
win counts as $1$ point, while a draw $0.5$ points. The average
score obtained in this way against SWH constitutes the \emph{Othello
League performance measure}, which we incorporate in this paper.

\subsection{N-tuple Network}

\begin{figure}
\centering{}\includegraphics{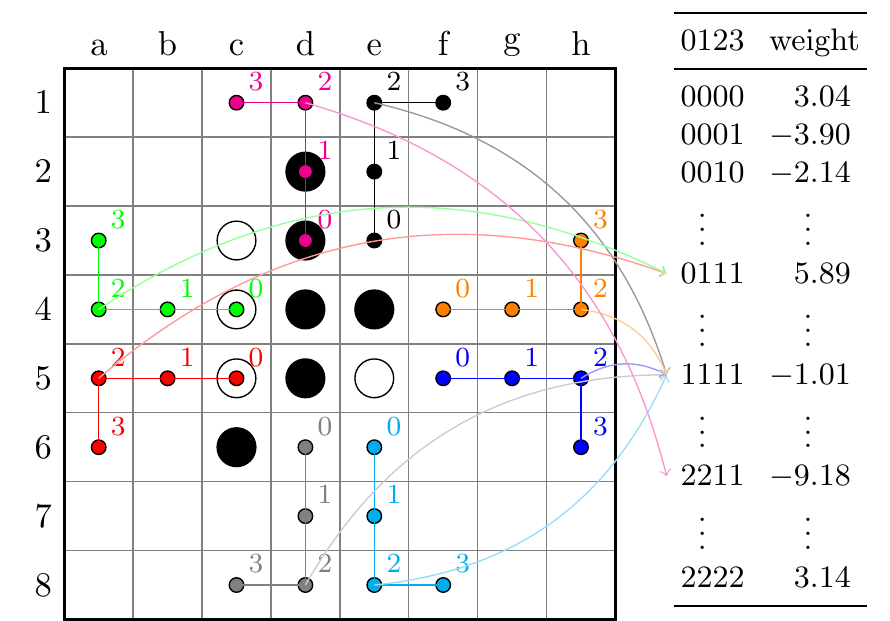} \protect\caption{\label{fig:OthelloNtuples}An $4$-tuple employed eight times to take
advantage of board symmetries (symmetric sampling). The eight symmetric
expansions of the $4$-tuple return, in total, $5\times-1.01+2\times5.89-9.18=-2.45$
for the given board state.}
\end{figure}

The best performing evaluation function in the Othello League is \emph{n-tuple
network} \cite{Samothrakis2012Coevolving}. N-tuple networks have
been first applied to optical character recognition problem by Bledsoe
and Browning \cite{Bledsoe1959Pattern}. For games, it have been used
first by Buro under the name of \emph{tabular value functions} \cite{Buro1997EvalOthello},
and later popularized by Lucas \cite{Lucas2008Learning}. According
to Szubert et al. their main advantages of n-tuple networks ``include
conceptual simplicity, speed of operation, and capability of realizing
nonlinear mappings to spaces of higher dimensionality'' \cite{Szubert2013Scalability}.

\begin{figure*}
\begin{centering}
\subfloat[\label{fig:RandomTuples}rand-$8\times4$ network consisting of $8$
randomly generated snake-shaped $4$-tuples ($648$ weights).]{\begin{centering}
\includegraphics{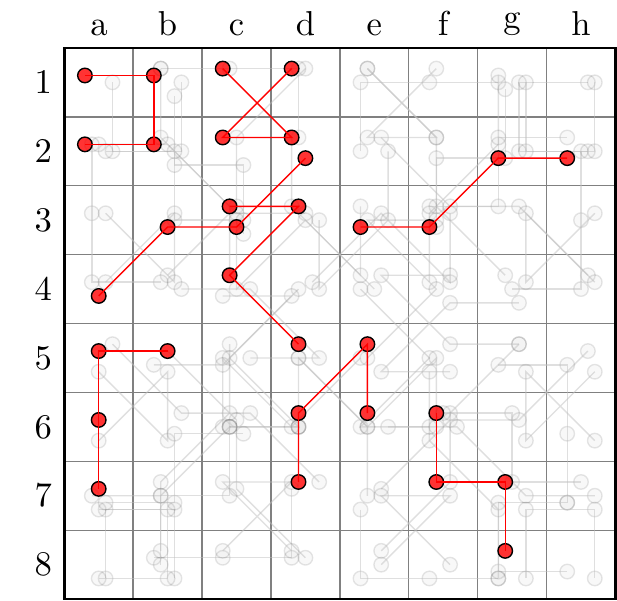}
\par\end{centering}

}~~~~~~~~~~~~~~~\subfloat[\label{fig:SystematicNTuples}all-$3$ network consisting of all $24$
straight $3$-tuples ($648$ weights).]{\begin{centering}
\includegraphics{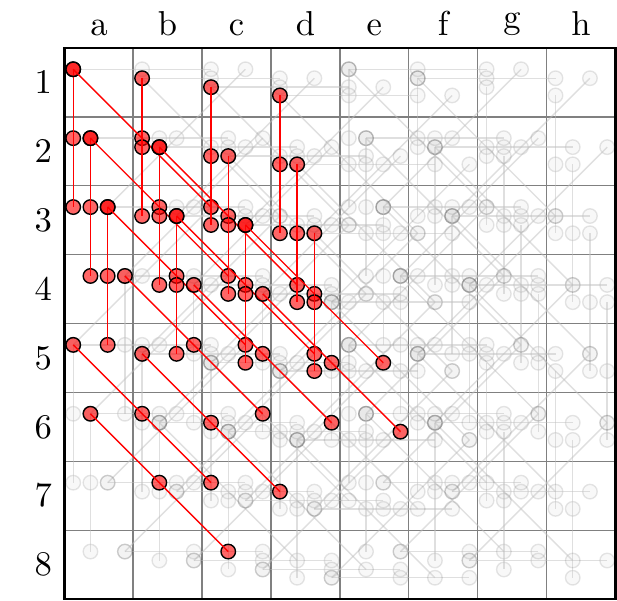}
\par\end{centering}

}
\par\end{centering}

\protect\caption{Comparison of rand-{*} and all-{*} n-tuple network architectures.
``Main'' n-tuples have been shown by red, while their symmetric
expansions by light gray. }
\end{figure*}
N-tuple network consists of $m$ $n_{i}$-tuples, where $n_{i}$ is
tuple's size. For a given board position $\mathbf{b}$, it returns
the sum of values returned by the individual n-tuples. The $i$th
$n_{i}$-tuple, for $i=1\dots m$, consists of a predetermined sequence
of board locations $(loc_{ij})_{j=1\dots n_{i}}$, and a look up table
$\text{LUT}_{i}$. The latter contains values for each board pattern
that can be observed on the sequence of board locations. Thus, n-tuple
network is a function 
\[
f(\mathbf{b})=\sum_{i=1}^{m}f_{i}(\mathbf{b})=\sum_{i=1}^{m}\text{LUT}_{i}\left[\text{index}\left(b_{loc_{i1}},\dots,b_{loc_{in_{i}}}\right)\right].
\]
Among possible ways to map the sequence to an $\text{index}$ in the
look up table, the following one is arguably convenient and computationally
efficient:
\[
\text{index}(\mathbf{v})=\sum_{j=1}^{|\mathbf{v}|}v_{j}c^{j-1},
\]
where $c$ is a constant denoting the number of possible values on
a single board square, and $\mathbf{v}$ is the sequence of board
values (the observed pattern) such that $0\le v_{j}<c$ for $j=1\dots|\mathbf{v}|$.
In the case of Othello, $c=3$, and white, empty, and black squares
are encoded as $0$, $1$, and $2$, respectively. 

The effectiveness of n-tuple networks is improved by using \emph{symmetric
sampling}, which exploits the inherent symmetries of the Othello board
\cite{lucas2007learning}. In symmetric sampling, a single n-tuple
is employed eight times, returning one value for each possible board
rotation and reflection. See Fig. \ref{fig:OthelloNtuples} for an
illustration.

\subsection{N-tuple Network Architecture}

Due to the spatial nature of game boards, n-tuples are usually consecutive
snake-shaped sequences of locations, although this is not a formal
requirement. If each n-tuple in a network is of the same size, we
denote it as $m\times n$-tuple network, having $m\times3^{n}$ weights.
Apart from choosing $n$ and $m$, an important design issue of n-tuples
network architecture is the location of individual n-tuples on the
board \cite{Szubert2013Scalability}.

\subsubsection{Random Snake-shaped N-tuple Network}

Thus it is surprising that so many investigations in game strategy
learning have involved\emph{ randomly generated} \emph{snake-shaped
n-tuple networks}. Lucas \cite{Lucas2008Learning} generated individual
n-tuples by starting from a random board location, then taking a random
walk of $6$ steps in any of the eight orthogonal or diagonal directions.
The repeated locations were ignored, thus the resulting n-tuples were
from $2$ to $6$ squares long. The same method Krawiec and Szubert
used for generating $7\times4$, $9\times5$ and $12\times6$-tuple
networks \cite{Krawiec2011Learning,Szubert2013Scalability}, and Thill
et al. \cite{Thill2012Reinforcement} for generating $70\times8$
tuple networks playing Connect Four.

An $m\times n$-tuple network generated in this way we will denote
as \emph{rand-}$m\times n$ (see Fig.~\ref{fig:RandomTuples} for
an example).

\subsubsection{Systematic Straight N-tuple Network}

Alternatively, we propose a deterministic method of constructing n-tuple
networks. Our \emph{systematic straight n-tuple networks} consist
of all possible vertical, horizontal or diagonal n-tuples placed on
the board. Its smallest representative is a network of $1$-tuples.
Thanks to symmetric sampling, only $10$ of them is required for an
$8\times8$ Othello board, and such $10\times1$-tuple network, which
we denote as all-$1$ contains $3\times10^{1}=30$ weights. all-$2$
network containing $32$ $2$-tuples is shown in Fig.~\ref{fig:SystematicNTuples}.
Table \ref{tab:Weights} shows the number of weights in selected architectures
of rand-{*} and all-{*} networks.

\subsubsection{Other Approaches}

Logistello~\cite{Buro2000OthelloMultiProbCut}, computer player,
which beat the human Othello world champion in 1997, used $11$ $n$-tuples
of $n\in\{3,10\}$, hand-crafted by an expert. External knowledge
has also been used by Manning~\cite{Manning2012NashOthello}, who,
generated a diverse $12\times6$-tuple network using random inputs
method from Breiman's Random Forests basing on a set of \num{10000}
labeled random games.

\subsection{Learning to Play Both Sides}

When a single player defined by its evaluation function is meant to
play both as black and white, it must interpret the result of the
evaluation function complementary depending on the color it plays.
There are three methods serving this purpose.

The first one is \emph{doubled function} (e.g., \cite{Thill2012Reinforcement}),
which simply employs two separate functions: one for playing white
and the other for playing black. It allows to fully separate the strategy
for white and black players. However, its disadvantage consists in
that two times more weights must be learned, and the experience learned
when playing as black does not used when playing as white and vice
versa.

\emph{Output negation} and \emph{board inversion }(e.g.,\emph{ }\cite{Runarsson2014Preference})
are alternatives to doubled function. They use only single set of
weights, reducing the search space and allowing to transfer the experience
between the white and black player. When using output negation, black
selects the move leading to a position with the maximal value of the
evaluation function whereas white selects the move leading to a position
with the minimal value.

If a player uses b\emph{oard inversion }it learns only to play black.
As the best black move it selects the one leading to the position
with the maximum value. If it has to play white, it temporarily flips
all the pieces on the board, so it can interpret the board as if it
played black. Then it selects the best `black' move, flips all the
pieces back, and plays the white piece in the selected location.

The \noun{swh} player uses output negation.

\section{Experiments and Results}

\begin{table}
\sisetup{table-number-alignment=center, table-format=4}
\def\arraystretch{1.1}

\begin{centering}
\begin{tabular}{lSllS}
\toprule 
architecture & \multicolumn{1}{c}{weights} &  & architecture & \multicolumn{1}{c}{weights}\tabularnewline
\midrule
all-$2$ ($32\times2$) & 288 &  & rand-$10\times3$ & 270\tabularnewline
all-$3$ ($24\times3$) & 648 &  & rand-$8\times4$ & 648\tabularnewline
all-$4$ ($21\times4$) & 1701 &  & rand-$7\times5$ & 1701\tabularnewline
\bottomrule
\end{tabular}
\par\end{centering}

\protect\caption{\label{tab:Weights}The number of weights for three pairs of \emph{systematic
straight} (all-{*}) and \emph{random snake-shaped }(rand-{*}) n-tuple
networks architectures.}
\end{table}

\subsection{Common Settings}

\subsubsection{Evolutionary Setup}

In order to compare different n-tuple network architectures, we performed
several computational experiments. In each of them the weights of
n-tuple networks have been learned by $(10+90)$ evolution strategy
\cite{beyer2002evolution} for $5000$ generations. The weights of
individuals in the initial population were drawn from the $[-0.1,0.1]$
interval. Evolution strategy used Gaussian mutation with $\sigma=1.0$.
The individual's fitness was calculated using the Othello League performance
measure estimated over $1000$ double games (cf. \ref{sub:Othello-League}). 

In total, $10^{10}$ games were played in each evolutionary run. This
makes our experiments exceptionally large compared to the previous
studies. For example, in a recent study concerning n-tuple networks
\cite{Szubert2013Scalability} $3\times10^{6}$ games were played.
Also, despite using the much simpler WPC representation, Samothrakis
et al. \cite{Samothrakis2012Coevolving} performed $10^{8}$ games
per run.

Such extensive experiment was possible due to efficient n-tuple network
and Othello implementation in Java, which is capable of running about
$1000$ games per second on a single CPU core. Thanks to it, we were
able to finish one evolutionary run in $28$ hours on a $6$-core
Intel(R) Core(TM) i7-2600 CPU @$3.40$GHz.

\subsubsection{Performance Evaluation}

We repeated each evolutionary $10$ times. Every $10$ generations,
we measured the (Othello League) performance of the fittest individual
in the population using \num{50000} double games. The performance
of the fittest individual from the last generation is identified with
method's performance. Since, the sample size is only $10$ per method,
for statistical analysis of the following experiments, we used non-parametric
Wilcoxon rank sum test (a.k.a. the Mann-Whitney U test) with the significance
level $\alpha=0.01$ and Holm's correction when comparing more than
two methods at once.

\subsection{Preliminary: Board Inversion vs. Output Negation}

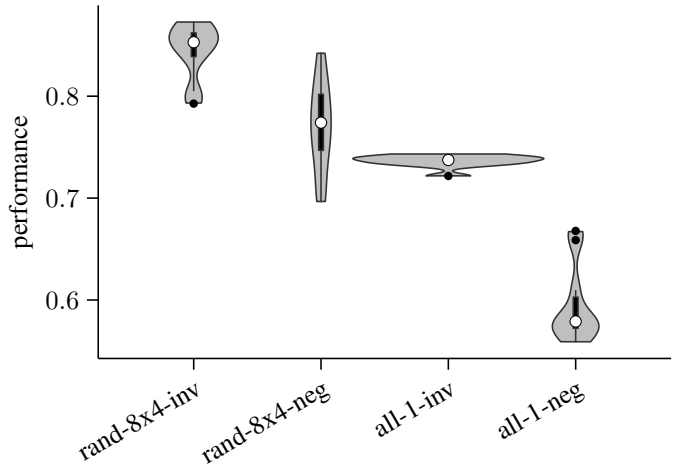
\begin{figure}
\begin{centering}
\input{img/inv-vs-neg.tex}
\par\end{centering}

\protect\caption{\label{fig:inv-vs.-neg}Comparison of output negation against board
inversion for two n-tuples architectures. The performance is measured
as the average score obtained against the Standard WPC Heuristic Player
at $\epsilon=0.1$ (Othello League performance). In each violin shape,
the white dot marks the median, the black boxes range from the lower
to the upper quartile, while the thin black lines represent $1.5$
interquartile range. Outliers beyond this range are denoted by black
dots. The outer shape shows the probability density of the data. }
\end{figure}

Figure \ref{fig:inv-vs.-neg} presents the results of learning with
board inversion against output negation for representatives of both
types of n-tuple networks architectures: rand-$8\times4$ having $8\times4^{3}=648$,
and all-$1$ with $10\times3^{1}=30$ weights.

The figure shows that board inversion surpasses output negation regardless
of the player architecture, which confirms a previous study of the
two methods for preference learning \cite{Runarsson2014Preference}.
The differences between the methods are statistically significant
(see also the detailed results in Table \ref{tab:Detailed}).

Moreover, visual inspection of the violin plots reveals that board
inversion leads to more robust learning, since the variance of performances
is lower. Therefore, in the following experiments we employ exclusively
board inversion.

\subsection{All Short Straight vs. Random Long Snake-shaped N-tuples}

In the main experiment, we compare n-tuple networks consisting of
all possible short straight n-tuples (all-$2$, all-$3$, and all-$4$)
with long random snake-shaped ones (rand-$10\times3$, rand-$8\times4$
and rand-$7\times5$). We chosen the number of n-tuples and size of
them to make the number of weights in of corresponding architectures
are equal, or, if impossible, similar (see Table \ref{tab:Weights}). 

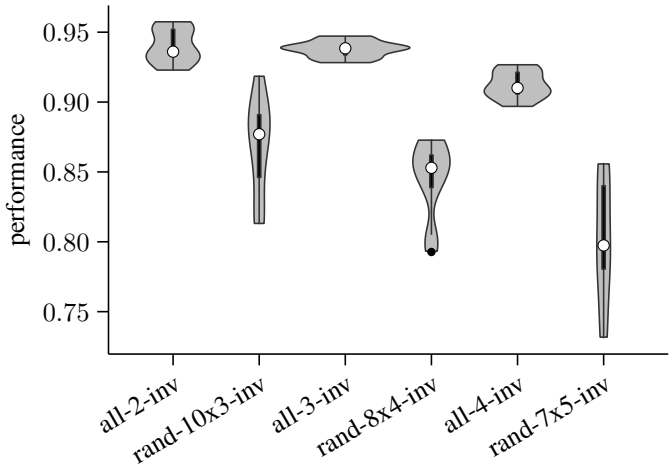
\begin{figure}
\begin{centering}
\input{img/all-vs-rand.tex}
\par\end{centering}

\protect\caption{\label{fig:all-vs.-rand}The comparison of all short straight n-tuple
networks (all-{*}) with random long snake-shaped n-tuple networks
(rand-{*}). The distribution of performances is presented as violin
plots (see Fig. \ref{fig:inv-vs.-neg} for explanation).}
\end{figure}

The results of the experiment are shown in Figure \ref{fig:all-vs.-rand}
as violin plots. Statistical analysis of three pairs having equal
or similar number of weights reveals that:
\begin{itemize}
\item all-$2$ is better than rand-$10\times3$,
\item all-$3$ is better than rand-$8\times4$, and
\item all-$4$ is better than rand-$7\times5$.
\end{itemize}
Let us notice that the differences in performance are substantial:
for the pair all-$2$ vs. rand-$10\times3$, where the difference
in performance is the lowest, the best result obtained by rand-$10\times3$
is still lower than the worst result obtained by all-$2$ (see Table
\ref{tab:Detailed} for details). 

All-{*} architectures are also more robust, due to lower variances
than rand-{*} architectures (cf. Fig. \ref{fig:all-vs.-rand}). This
is because the variance of rand-{*} architectures is attributed to
both its random initialization and non-deterministic learning process,
while the variance of all-{*} is only due to the latter.

\subsection{2-tuples are Long Enough}

Intuitively, longer n-tuples should lead to higher network's performance,
since they can `react' to patterns that the shorter ones cannot.
However, the results presented in Fig.~\ref{fig:all-vs.-rand} show
no evidence that this is a case. Despite having two times more weights,
all-$3$ does not provide better performance than all-$2$ (no statistical
difference). Furthermore, all-$4$ is significantly worse than both
than all-$2$ and all-$3$.

\begin{figure*}
\begin{centering}
\includegraphics{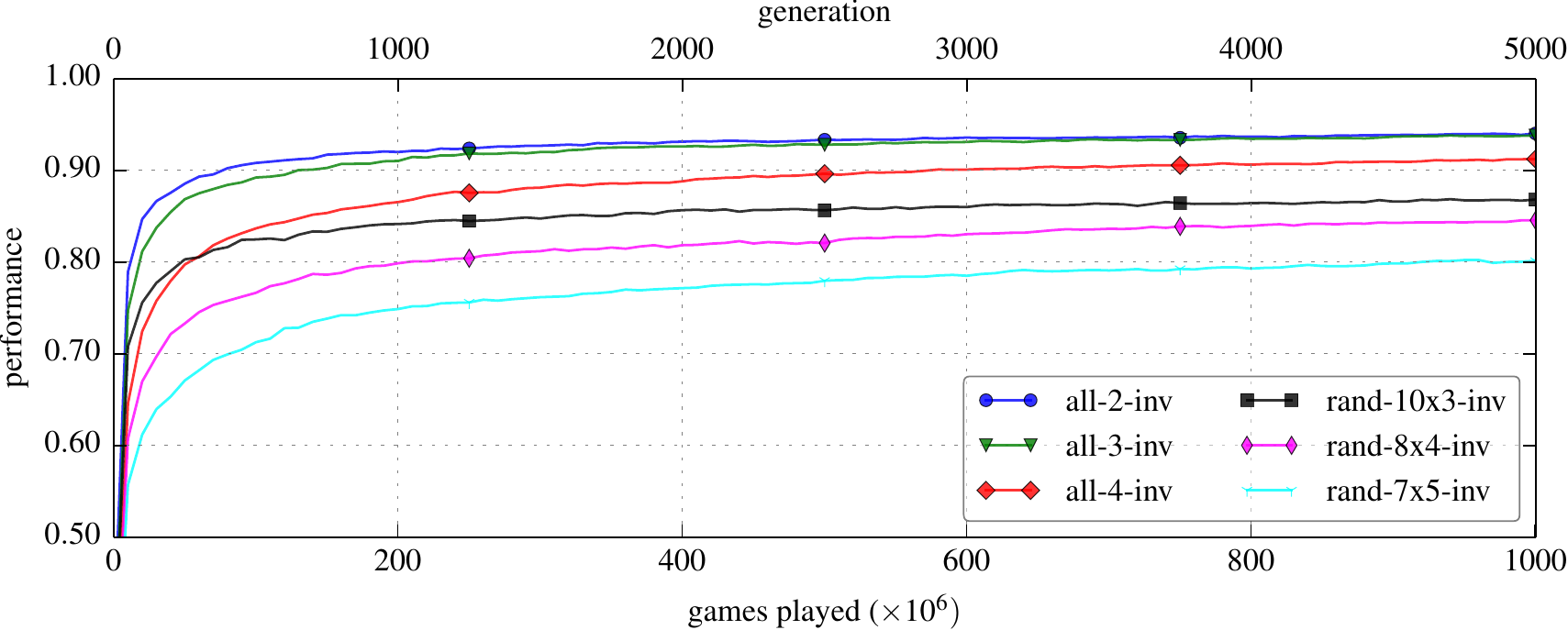}
\par\end{centering}

\protect\caption{\label{fig:LearningRate}Pace of learning of six analyzed n-tuple
networks architectures. Each point on the plot denotes the average
performance of method's fittest individual in a given generation.}
\end{figure*}

Figure~\ref{fig:LearningRate} shows the pace of learning for each
of six analyzed architectures. It plots methods' performance as a
function of computational effort, which is proportional to the number
of generations. 

The figure suggests that all-$2$ is not only the best (together with
all-$3$) in the long run, but it is also the method that learns the
quickest. all-$3$ catches up all-$2$ eventually, but it does not
seem to be able to surpass it. all-$4$ learns even slower than all-$3$.
Although the gap between all-$3$ and all-$4$ decreases over time,
it is still noticeable after $5000$ generations.

Thus, our results suggest that for Othello, all-$2$ with just $288$
weights, the smallest among the six considered n-tuple network architectures,
is also the best one.

\subsection{Othello League Results}

The best player obtained in this research consists of all $2$-tuples;
its performance is $0.9592$ with $95\%$ confidence delta of $\pm0.0012$.
This result is significantly higher than the best results reported
to this date in the Othello League (see \ref{tab:OthelloLeague}).
Notice also how small it is (in terms of the number of weights) compared
to other players in the league. Unfortunately, the on-line Othello
League accepts only players employing output negation; it does not
allow for board inversion. Thus, our player could not be submitted
to the Othello League.

\begin{table}
\def\arraystretch{1.3}
\renewcommand{\tabcolsep}{0.1cm}
\rowcolors{3}{lightgray}{}

\begin{centering}
\begin{tabular}{lllS[table-format=5.0]S[table-format=1.4]}
\toprule 
\multirow{1}{*}{date} & \multirow{1}{*}{player name} & \multirow{1}{*}{encoding} & \multicolumn{1}{l}{weights} & \multicolumn{1}{c}{performance}\tabularnewline
\midrule
\textbf{n/a} & \textbf{all-$2$-inv} & \textbf{n-tuple network} & \textbf{288} & \textbf{0.9592}\tabularnewline
\textbf{2013-09-17} & \textbf{wj-1-2-3-tuples} & \textbf{n-tuple network} & \textbf{966} & \textbf{0.9149}\tabularnewline
2011-01-30 & epTDLmpx\_12x6 \cite{Szubert2013Scalability} & n-tuple network & 3240 & 0.871\tabularnewline
2011-01-28 & prb\_nt15\_001 & n-tuple network & 6561 & 0.845\tabularnewline
2011-01-25 & epTDLxover \cite{Szubert2013Scalability} & n-tuple network & 4698 & 0.83\tabularnewline
2008-05-03 & t15x6x8 & n-tuple network & 10935 & 0.805\tabularnewline
2008-05-03 & x30x6x8 & n-tuple network & 21870 & 0.73\tabularnewline
2008-03-28 & Stunner & n-tuple network & 7725 & 0.675\tabularnewline
2007-09-14 & MLP(...)312-ties0.FF & neural network & 1915 & 0.62\tabularnewline
\bottomrule
\end{tabular}
\par\end{centering}

\protect\caption{\label{tab:OthelloLeague}Selected milestones (improvements) in the
on-line Othello Position Evaluation Function League since September
2007. The table consists also all-$2$, not submitted to the League
since it uses board inversion. The performances of all but the three
best players come from the Othello League website and have been estimated
using $50$ double games. The performances of all-2 and wj-1-2-3-tuples
players have been estimated using \num{50000} double games, and the
performance of epTDLmpx\_12x6 has been reported in \cite{Szubert2013Scalability}.}
\end{table}

To be accepted in the Othello League, we performed some experiments
also with output negation. The best output negation player we were
able to evolve was submitted under the name of wj-1-2-3-tuples. It
consists of all straight 1-, 2-, and 3-tuples, thus having $966$
weights in total.

wj-1-2-3-tuples took the lead in the league and is the first player
exceeding the performance of $0.9$. It obtained $0.94$ in the league,
but this result should be taken with care, since to evaluate player's
performance Othello League plays just $100$ games. We estimate its
performance to $0.9149\pm0.0017$ basing on \num{50000} double games. 

We suspect that the performance of ca. $0.96$ against Standard WPC
Heuristic player that all-$2$ and all-$3$ converge to, cannot be
significantly improved at $1$-ply. $\epsilon=0.1$ random moves using
in all games leads to the situation when even a perfect-playing player
cannot guarantee not losing a game.

Despite the first place obtained in the Othello League, the evolved
player is not good in `general', against a variety of opponents,
because is was evolved specifically to play against the Standard WPC
Heuristic player. When evaluated against random WPC players (the expected
utility measure \cite{Chong2012improving,jaskowski2013improving}),
the best all-$2$ player obtains a score of only $0.9584\pm0.0012$.
This is not much, since with considerably less computational effort
that used in this paper, it is possible to evolve an n-tuple player
scoring $>0.99$ \cite{Liskowski2012Msc,Szubert2013Scalability}.
However, our goal here was not to design good players in general,
but to compare different position evaluation functions.

The best all-$2$ player evolved in this paper is printed in Fig.~\ref{fig:all-2}.

\section{Discussion: the more weights, the worse for evolution?}

We have shown that among all-{*} methods, the more weights the worse
results; the same applies to rand-{*} methods (see Fig. \ref{fig:all-vs.-rand}).
This finding confirms the one of Szubert et al. \cite{Szubert2013Scalability},
who found out that among the networks of rand-$12\times6$ ($8748$
weights), rand-$9\times5$ ($2187$ weights), and rand-$7\times4$
($567$ weights), it is the latter that allows (co)evolutionary algorithm
for obtaining best results. The authors stated that this effect it
due to the higher dimensionality of the search space, for which ``the
weight mutation operator is not sufficiently efficient to elaborate
fast progress''. 

Although we do not challenge this claim, our results suggest that
the number of weights in a network is not the only performance factor.
all-$4$ has $1701$ weights, thus, the dimensionality of its search
space is considerably higher than the one for rand-$10\times3$ and
rand-$8\times4$, which have $270$ and $648$ weights, respectively.
Nonetheless, among these three architectures, it is the all-$4$ network
that obtains the highest performance (see Fig. \ref{fig:all-vs.-rand}).
Therefore, the second performance factor in learning an n-tuple network
is its (proper or not) architecture.

Finally, let us notice that an alternative to a fixed n-tuple network
architecture is a self-adaptive one, which can change in response
to variation operators \cite{Szubert2013Scalability}, such as mutation
or crossover. Although such architecture is, in principle, more flexible,
it adds another dimension to the search space, making the learning
problem even harder.

\section{Conclusions}

In this paper, we have analyzed n-tuple network architectures for
position evaluation function in board games. We have shown that a
network consisting of all possible, systematically generated, short
n-tuples leads to a significantly better play than long random snake-shaped
tuples originally used by Lucas \cite{lucas2007learning}. With a
simple network consisting of all possible straight $2$-tuples (with
just $288$ weights) we were able to beat the best result in the on-line
Othello League (having usually many times more weights). 

Moreover, our results suggest that tuples longer than $2$ give no
advantage, causing slower learning rate, at the same time. This is
surprising, since capturing opponent's pieces in Othello requires
a line of at least three pieces (e.g. white, black, white).

Let us emphasize that our result has been obtained in an intensive
computational experiment involving $5000$ generations, an order of
magnitude more than other studies in this domain. Nevertheless, it
remains to be seen whether they hold for different experimental settings.
We used evolution against an expert player in $1$-ply $\epsilon$-Othello.
The interesting questions are: i) whether our systematic short $2$-tuple
network is also advantageous for reinforcement learning, such as temporal
difference learning, and ii) whether such networks are also profitable
for other board games, e.g. Connect Four.

\section*{Acknowledgment }

This work has been supported by the Polish National Science Centre
grant no. DEC-2013/09/D/ST6/03932. The computations have been performed
in Pozna\'{n} Supercomputing and Networking Center. The author would
like to thank Marcin Szubert for his helpful remarks on an earlier
version of this article.

\begin{table*}[t]
\sisetup{table-number-alignment = center, round-mode=places, round-precision=4, table-format=1.4}
\rowcolors{2}{lightgray}{}
\centering
\begin{tabular}{lSSSSSSSSSSSS}
\toprule
{} &     {mean} &   {median} &       0 &       1 &       2 &       3 &       4 &       5 &       6 &       7 &       8 &       9 \\
\midrule
all-2-inv     &  0.94010 &  0.93605 &  0.9529 &  0.9330 &  0.9270 &  0.9578 &  0.9327 &  0.9349 &  0.9225 &  0.9372 &  0.9495 &  0.9535 \\
all-3-inv     &  0.93791 &  0.93845 &  0.9326 &  0.9314 &  0.9402 &  0.9282 &  0.9382 &  0.9387 &  0.9404 &  0.9472 &  0.9452 &  0.9370 \\
all-4-inv     &  0.91223 &  0.91010 &  0.9108 &  0.8967 &  0.9171 &  0.9076 &  0.9094 &  0.9230 &  0.9051 &  0.9030 &  0.9269 &  0.9227 \\
rand-10x3-inv &  0.86806 &  0.87700 &  0.8649 &  0.8145 &  0.8834 &  0.8920 &  0.8126 &  0.8398 &  0.9190 &  0.8958 &  0.8706 &  0.8880 \\
rand-8x4-inv  &  0.84541 &  0.85295 &  0.8519 &  0.7927 &  0.8050 &  0.8595 &  0.8366 &  0.8630 &  0.8728 &  0.8733 &  0.8540 &  0.8453 \\
rand-7x5-inv  &  0.80118 &  0.79740 &  0.8285 &  0.8565 &  0.7978 &  0.8559 &  0.7398 &  0.7310 &  0.7803 &  0.7810 &  0.7970 &  0.8440 \\
rand-8x4-neg  &  0.76978 &  0.77405 &  0.8104 &  0.6959 &  0.8429 &  0.7133 &  0.7453 &  0.7817 &  0.8085 &  0.7517 &  0.7668 &  0.7813 \\
all-1-inv     &  0.73621 &  0.73740 &  0.7337 &  0.7395 &  0.7423 &  0.7355 &  0.7434 &  0.7217 &  0.7317 &  0.7395 &  0.7361 &  0.7387 \\
all-1-neg     &  0.59462 &  0.57895 &  0.5799 &  0.5584 &  0.5821 &  0.5650 &  0.6588 &  0.5780 &  0.6678 &  0.6097 &  0.5713 &  0.5752 \\
\bottomrule
\end{tabular}

\protect\caption{\label{tab:Detailed}Performances obtained in $10$ evolutionary runs
of all n-tuple network architectures considered in this study. Each
value is an average score in \num{50000} double games against Standard
WPC Heuristic in $\epsilon$-Othello, where $\epsilon=0.1$.}
\end{table*}

\begin{figure}
{\scriptsize{}}
\begin{lstlisting}[basicstyle={\tiny}]
{ 32
  { 2 8 { 6 7 } { 55 63 } { 7 15 } { 56 57 } { 62 63 } { 0 1 } { 48 56 } { 0 8 } 
    { 57.64 -91.70 111.05 -82.30 74.42 -96.30 211.47 -53.91 142.45 }  }
  { 2 4 { 49 56 } { 0 9 } { 7 14 } { 54 63 } 
    { 84.51 -33.37 -29.83 -72.21 -199.31 -18.72 -98.04 22.34 185.84 }  }
  { 2 8 { 1 2 } { 15 23 } { 8 16 } { 40 48 } { 57 58 } { 5 6 } { 47 55 } { 61 62 } 
    { -42.97 109.76 -66.10 84.67 158.09 -148.21 -11.94 -94.92 111.52 }  }
  { 2 8 { 48 49 } { 54 55 } { 8 9 } { 54 62 } { 6 14 } { 14 15 } { 49 57 } { 1 9 } 
    { -144.66 16.16 -78.97 -83.93 -7.37 -15.97 -102.98 -51.68 -3.99 }  }
  { 2 8 { 41 48 } { 8 17 } { 6 13 } { 50 57 } { 53 62 } { 1 10 } { 46 55 } { 15 22 } 
    { -115.74 -43.49 -117.28 33.32 -14.29 12.85 -18.33 2.95 91.17 }  }
  { 2 8 { 4 5 } { 16 24 } { 39 47 } { 58 59 } { 60 61 } { 32 40 } { 2 3 } { 23 31 } 
    { 31.28 -32.65 37.46 -20.60 273.39 -90.25 60.16 -151.55 -30.45 }  }
  { 2 8 { 2 10 } { 16 17 } { 46 47 } { 22 23 } { 50 58 } { 40 41 } { 5 13 } { 53 61 } 
    { 93.18 -27.12 -16.53 -2.52 -80.43 62.23 36.46 39.86 -97.27 }  }
  { 2 8 { 52 61 } { 23 30 } { 51 58 } { 16 25 } { 5 12 } { 2 11 } { 38 47 } { 33 40 } 
    { -42.75 36.91 44.65 -49.19 -19.01 -55.54 10.60 -36.50 74.37 }  }
  { 2 4 { 24 32 } { 59 60 } { 3 4 } { 31 39 } 
    { 43.77 -122.45 55.43 -5.38 284.39 -103.51 79.44 -94.92 157.02 }  }
  { 2 8 { 52 60 } { 24 25 } { 51 59 } { 32 33 } { 4 12 } { 30 31 } { 38 39 } { 3 11 } 
    { 99.45 20.89 -10.97 15.39 -28.77 16.65 -16.89 -12.47 -18.57 }  }
  { 2 8 { 30 39 } { 25 32 } { 3 12 } { 51 60 } { 24 33 } { 31 38 } { 52 59 } { 4 11 } 
    { 14.85 88.04 -3.28 26.88 33.46 -19.67 -5.65 28.85 -43.03 }  }
  { 2 8 { 39 46 } { 53 60 } { 4 13 } { 17 24 } { 50 59 } { 32 41 } { 3 10 } { 22 31 } 
    { -32.77 22.27 -51.36 -2.01 -65.96 -33.23 16.39 8.59 -28.07 }  }
  { 2 8 { 47 54 } { 5 14 } { 40 49 } { 49 58 } { 2 9 } { 14 23 } { 9 16 } { 54 61 } 
    { -82.49 -5.56 10.19 -68.91 29.84 -37.59 -69.56 -0.20 10.63 }  }
  { 2 4 { 6 15 } { 48 57 } { 55 62 } { 1 8 } 
    { -34.15 20.89 -135.36 79.22 30.55 20.13 35.27 -11.57 16.43 }  }
  { 2 8 { 13 14 } { 9 10 } { 41 49 } { 9 17 } { 53 54 } { 46 54 } { 49 50 } { 14 22 } 
    { 20.98 -44.43 30.94 -64.79 -27.71 -37.59 17.05 4.34 -17.03 }  }
  { 2 4 { 45 54 } { 9 18 } { 14 21 } { 42 49 } 
    { 43.08 104.43 14.69 24.49 31.96 -14.64 -51.11 -22.12 14.48 }  }
  { 2 8 { 38 46 } { 12 13 } { 52 53 } { 22 30 } { 50 51 } { 10 11 } { 33 41 } { 17 25 } 
    { 19.75 39.63 -16.15 1.75 -38.84 9.21 6.77 14.85 19.99 }  }
  { 2 8 { 41 42 } { 13 21 } { 45 46 } { 45 53 } { 10 18 } { 21 22 } { 17 18 } { 42 50 } 
    { 5.28 -38.02 12.64 -90.60 60.60 5.96 60.38 27.61 3.00 }  }
  { 2 8 { 37 46 } { 10 19 } { 17 26 } { 43 50 } { 22 29 } { 44 53 } { 13 20 } { 34 41 } 
    { -13.44 19.48 -13.49 0.72 -59.65 -3.23 45.27 45.31 30.39 }  }
  { 2 4 { 25 33 } { 30 38 } { 11 12 } { 51 52 } 
    { 8.66 14.83 15.73 -34.15 32.08 -9.15 15.15 41.61 66.03 }  }
  { 2 8 { 25 26 } { 12 20 } { 11 19 } { 33 34 } { 44 52 } { 43 51 } { 37 38 } { 29 30 } 
    { -71.70 12.07 -54.50 18.12 86.36 22.27 -56.07 -4.46 -43.54 }  }
  { 2 8 { 30 37 } { 43 52 } { 29 38 } { 26 33 } { 44 51 } { 25 34 } { 11 20 } { 12 19 } 
    { 11.34 25.64 -28.34 41.82 73.33 -26.18 -0.64 -25.88 -29.12 }  }
  { 2 8 { 42 51 } { 21 30 } { 45 52 } { 11 18 } { 12 21 } { 18 25 } { 38 45 } { 33 42 } 
    { -32.37 28.60 13.65 -48.41 -13.25 -63.15 -30.60 18.99 22.64 }  }
  { 2 4 { 10 17 } { 13 22 } { 46 53 } { 41 50 } 
    { 31.89 -78.88 -32.75 44.88 -42.65 39.91 26.48 -12.34 -46.59 }  }
  { 2 8 { 34 42 } { 21 29 } { 20 21 } { 18 26 } { 18 19 } { 42 43 } { 37 45 } { 44 45 } 
    { 0.66 -29.13 12.95 -17.71 -71.59 11.40 31.52 -4.66 79.89 }  }
  { 2 4 { 35 42 } { 18 27 } { 21 28 } { 36 45 } 
    { 76.53 141.05 -5.04 61.89 16.13 43.95 -4.87 182.85 -92.46 }  }
  { 2 4 { 26 34 } { 43 44 } { 29 37 } { 19 20 } 
    { -10.60 11.60 -8.56 -6.06 -137.99 -8.81 -3.62 3.30 -39.96 }  }
  { 2 8 { 20 28 } { 34 35 } { 28 29 } { 36 37 } { 26 27 } { 35 43 } { 36 44 } { 19 27 } 
    { 38.03 56.90 -26.64 -61.84 21.69 -116.98 7.91 53.48 -58.83 }  }
  { 2 8 { 19 28 } { 36 43 } { 29 36 } { 20 27 } { 26 35 } { 27 34 } { 35 44 } { 28 37 } 
    { -17.67 81.78 -63.71 15.62 2.16 16.63 -14.70 43.61 -24.67 }  }
  { 2 4 { 34 43 } { 37 44 } { 20 29 } { 19 26 } 
    { -18.74 -52.99 -3.10 -31.68 -181.22 -24.62 32.63 58.19 40.79 }  }
  { 2 4 { 35 36 } { 27 28 } { 28 36 } { 27 35 } 
    { -100.00 -28.48 -4.99 -63.27 -80.05 -55.95 22.63 22.57 133.65 }  }
  { 2 2 { 28 35 } { 27 36 } 
    { 20.00 -52.52 -12.54 40.64 192.61 14.14 4.70 -1.81 -42.70 }  }
}
\end{lstlisting}
{\scriptsize \par}

\protect\caption{\label{fig:all-2}N-tuple network representing the best evolved all-$2$
player in the online Othello League format. The player contains $32$
$2$-tuples. Each one has at most $8$ symmetric expansions (sometimes
$4$ or $2$). The fields are numbered from $0$ to $63$ in row-wise
fashion. The network has $32\times9=288$ weights. The player uses
board inversion. Its Othello League performance is $0.9584\pm0.0012$.}
\end{figure}

\bibliographystyle{IEEEtran}
\addcontentsline{toc}{section}{\refname}\bibliography{all,library,wjaskowski}

\end{document}

%% file: img/inv-vs-neg.tex
\begin{tikzpicture}[x=1pt,y=1pt]
\definecolor[named]{fillColor}{rgb}{1.00,1.00,1.00}
\path[use as bounding box,fill=fillColor,fill opacity=0.00] (0,0) rectangle (251.50,179.23);
\begin{scope}
\path[clip] (  0.00,  0.00) rectangle (251.50,179.23);
\definecolor[named]{drawColor}{rgb}{1.00,1.00,1.00}
\definecolor[named]{fillColor}{rgb}{1.00,1.00,1.00}

\path[draw=drawColor,line width= 0.6pt,line join=round,line cap=round,fill=fillColor] (  0.00,  0.00) rectangle (251.50,179.23);
\end{scope}
\begin{scope}
\path[clip] ( 33.94, 44.73) rectangle (250.65,178.38);
\definecolor[named]{drawColor}{rgb}{0.20,0.20,0.20}
\definecolor[named]{fillColor}{rgb}{0.75,0.75,0.75}

\path[draw=drawColor,line width= 0.6pt,line join=round,line cap=round,fill=fillColor] ( 67.16,141.42) --
	( 67.09,141.69) --
	( 67.03,141.96) --
	( 66.97,142.23) --
	( 66.93,142.51) --
	( 66.90,142.78) --
	( 66.87,143.05) --
	( 66.86,143.32) --
	( 66.85,143.59) --
	( 66.86,143.86) --
	( 66.87,144.13) --
	( 66.90,144.41) --
	( 66.93,144.68) --
	( 66.97,144.95) --
	( 67.03,145.22) --
	( 67.09,145.49) --
	( 67.15,145.76) --
	( 67.23,146.03) --
	( 67.31,146.31) --
	( 67.40,146.58) --
	( 67.49,146.85) --
	( 67.59,147.12) --
	( 67.68,147.39) --
	( 67.79,147.66) --
	( 67.89,147.93) --
	( 67.99,148.20) --
	( 68.09,148.48) --
	( 68.19,148.75) --
	( 68.29,149.02) --
	( 68.38,149.29) --
	( 68.46,149.56) --
	( 68.54,149.83) --
	( 68.61,150.10) --
	( 68.67,150.38) --
	( 68.73,150.65) --
	( 68.77,150.92) --
	( 68.80,151.19) --
	( 68.82,151.46) --
	( 68.83,151.73) --
	( 68.82,152.00) --
	( 68.80,152.28) --
	( 68.77,152.55) --
	( 68.72,152.82) --
	( 68.67,153.09) --
	( 68.59,153.36) --
	( 68.51,153.63) --
	( 68.41,153.90) --
	( 68.30,154.18) --
	( 68.17,154.45) --
	( 68.04,154.72) --
	( 67.89,154.99) --
	( 67.74,155.26) --
	( 67.57,155.53) --
	( 67.39,155.80) --
	( 67.21,156.07) --
	( 67.01,156.35) --
	( 66.81,156.62) --
	( 66.61,156.89) --
	( 66.39,157.16) --
	( 66.18,157.43) --
	( 65.95,157.70) --
	( 65.73,157.97) --
	( 65.50,158.25) --
	( 65.27,158.52) --
	( 65.03,158.79) --
	( 64.80,159.06) --
	( 64.56,159.33) --
	( 64.33,159.60) --
	( 64.10,159.87) --
	( 63.86,160.15) --
	( 63.63,160.42) --
	( 63.41,160.69) --
	( 63.19,160.96) --
	( 62.97,161.23) --
	( 62.76,161.50) --
	( 62.55,161.77) --
	( 62.35,162.04) --
	( 62.16,162.32) --
	( 61.98,162.59) --
	( 61.81,162.86) --
	( 61.65,163.13) --
	( 61.50,163.40) --
	( 61.37,163.67) --
	( 61.24,163.94) --
	( 61.13,164.22) --
	( 61.04,164.49) --
	( 60.95,164.76) --
	( 60.89,165.03) --
	( 60.83,165.30) --
	( 60.80,165.57) --
	( 60.77,165.84) --
	( 60.77,166.12) --
	( 60.77,166.39) --
	( 60.79,166.66) --
	( 60.83,166.93) --
	( 60.88,167.20) --
	( 60.94,167.47) --
	( 61.01,167.74) --
	( 61.09,168.01) --
	( 61.19,168.29) --
	( 61.30,168.56) --
	( 61.42,168.83) --
	( 61.55,169.10) --
	( 61.69,169.37) --
	( 61.84,169.64) --
	( 62.00,169.91) --
	( 62.17,170.19) --
	( 62.35,170.46) --
	( 62.54,170.73) --
	( 62.73,171.00) --
	( 62.94,171.27) --
	( 63.15,171.54) --
	( 63.38,171.81) --
	( 63.61,172.09) --
	( 76.50,172.09) --
	( 76.73,171.81) --
	( 76.95,171.54) --
	( 77.17,171.27) --
	( 77.37,171.00) --
	( 77.57,170.73) --
	( 77.76,170.46) --
	( 77.94,170.19) --
	( 78.11,169.91) --
	( 78.27,169.64) --
	( 78.42,169.37) --
	( 78.56,169.10) --
	( 78.69,168.83) --
	( 78.81,168.56) --
	( 78.92,168.29) --
	( 79.01,168.01) --
	( 79.10,167.74) --
	( 79.17,167.47) --
	( 79.23,167.20) --
	( 79.28,166.93) --
	( 79.31,166.66) --
	( 79.34,166.39) --
	( 79.34,166.12) --
	( 79.33,165.84) --
	( 79.31,165.57) --
	( 79.27,165.30) --
	( 79.22,165.03) --
	( 79.15,164.76) --
	( 79.07,164.49) --
	( 78.98,164.22) --
	( 78.87,163.94) --
	( 78.74,163.67) --
	( 78.61,163.40) --
	( 78.46,163.13) --
	( 78.30,162.86) --
	( 78.13,162.59) --
	( 77.95,162.32) --
	( 77.76,162.04) --
	( 77.56,161.77) --
	( 77.35,161.50) --
	( 77.14,161.23) --
	( 76.92,160.96) --
	( 76.70,160.69) --
	( 76.47,160.42) --
	( 76.24,160.15) --
	( 76.01,159.87) --
	( 75.78,159.60) --
	( 75.54,159.33) --
	( 75.31,159.06) --
	( 75.07,158.79) --
	( 74.84,158.52) --
	( 74.61,158.25) --
	( 74.38,157.97) --
	( 74.15,157.70) --
	( 73.93,157.43) --
	( 73.71,157.16) --
	( 73.50,156.89) --
	( 73.30,156.62) --
	( 73.09,156.35) --
	( 72.90,156.07) --
	( 72.72,155.80) --
	( 72.54,155.53) --
	( 72.37,155.26) --
	( 72.22,154.99) --
	( 72.07,154.72) --
	( 71.93,154.45) --
	( 71.81,154.18) --
	( 71.70,153.90) --
	( 71.60,153.63) --
	( 71.51,153.36) --
	( 71.44,153.09) --
	( 71.38,152.82) --
	( 71.34,152.55) --
	( 71.31,152.28) --
	( 71.29,152.00) --
	( 71.28,151.73) --
	( 71.29,151.46) --
	( 71.31,151.19) --
	( 71.34,150.92) --
	( 71.38,150.65) --
	( 71.43,150.38) --
	( 71.50,150.10) --
	( 71.57,149.83) --
	( 71.65,149.56) --
	( 71.73,149.29) --
	( 71.82,149.02) --
	( 71.92,148.75) --
	( 72.02,148.48) --
	( 72.12,148.20) --
	( 72.22,147.93) --
	( 72.32,147.66) --
	( 72.42,147.39) --
	( 72.52,147.12) --
	( 72.62,146.85) --
	( 72.71,146.58) --
	( 72.80,146.31) --
	( 72.88,146.03) --
	( 72.95,145.76) --
	( 73.02,145.49) --
	( 73.08,145.22) --
	( 73.13,144.95) --
	( 73.18,144.68) --
	( 73.21,144.41) --
	( 73.23,144.13) --
	( 73.25,143.86) --
	( 73.25,143.59) --
	( 73.25,143.32) --
	( 73.23,143.05) --
	( 73.21,142.78) --
	( 73.18,142.51) --
	( 73.13,142.23) --
	( 73.08,141.96) --
	( 73.02,141.69) --
	( 72.95,141.42) --
	( 67.16,141.42) --
	cycle;

\path[draw=drawColor,line width= 0.6pt,line join=round,line cap=round,fill=fillColor] (116.59,104.16) --
	(116.55,104.71) --
	(116.50,105.27) --
	(116.46,105.82) --
	(116.42,106.38) --
	(116.38,106.93) --
	(116.34,107.49) --
	(116.30,108.05) --
	(116.27,108.60) --
	(116.23,109.16) --
	(116.20,109.71) --
	(116.17,110.27) --
	(116.13,110.82) --
	(116.10,111.38) --
	(116.07,111.93) --
	(116.03,112.49) --
	(116.00,113.05) --
	(115.96,113.60) --
	(115.93,114.16) --
	(115.89,114.71) --
	(115.85,115.27) --
	(115.80,115.82) --
	(115.76,116.38) --
	(115.71,116.94) --
	(115.67,117.49) --
	(115.62,118.05) --
	(115.56,118.60) --
	(115.51,119.16) --
	(115.45,119.71) --
	(115.40,120.27) --
	(115.34,120.82) --
	(115.28,121.38) --
	(115.22,121.94) --
	(115.16,122.49) --
	(115.10,123.05) --
	(115.04,123.60) --
	(114.98,124.16) --
	(114.93,124.71) --
	(114.87,125.27) --
	(114.82,125.83) --
	(114.77,126.38) --
	(114.72,126.94) --
	(114.67,127.49) --
	(114.63,128.05) --
	(114.59,128.60) --
	(114.55,129.16) --
	(114.52,129.72) --
	(114.49,130.27) --
	(114.46,130.83) --
	(114.44,131.38) --
	(114.42,131.94) --
	(114.41,132.49) --
	(114.40,133.05) --
	(114.39,133.60) --
	(114.39,134.16) --
	(114.39,134.72) --
	(114.40,135.27) --
	(114.41,135.83) --
	(114.42,136.38) --
	(114.44,136.94) --
	(114.46,137.49) --
	(114.48,138.05) --
	(114.50,138.61) --
	(114.53,139.16) --
	(114.56,139.72) --
	(114.60,140.27) --
	(114.64,140.83) --
	(114.67,141.38) --
	(114.72,141.94) --
	(114.76,142.49) --
	(114.81,143.05) --
	(114.85,143.61) --
	(114.90,144.16) --
	(114.96,144.72) --
	(115.01,145.27) --
	(115.06,145.83) --
	(115.12,146.38) --
	(115.18,146.94) --
	(115.24,147.50) --
	(115.30,148.05) --
	(115.36,148.61) --
	(115.42,149.16) --
	(115.48,149.72) --
	(115.55,150.27) --
	(115.61,150.83) --
	(115.68,151.38) --
	(115.74,151.94) --
	(115.81,152.50) --
	(115.87,153.05) --
	(115.94,153.61) --
	(116.00,154.16) --
	(116.07,154.72) --
	(116.14,155.27) --
	(116.20,155.83) --
	(116.27,156.39) --
	(116.33,156.94) --
	(116.39,157.50) --
	(116.46,158.05) --
	(116.52,158.61) --
	(116.59,159.16) --
	(116.65,159.72) --
	(116.71,160.27) --
	(119.71,160.27) --
	(119.78,159.72) --
	(119.84,159.16) --
	(119.90,158.61) --
	(119.97,158.05) --
	(120.03,157.50) --
	(120.09,156.94) --
	(120.16,156.39) --
	(120.22,155.83) --
	(120.29,155.27) --
	(120.35,154.72) --
	(120.42,154.16) --
	(120.49,153.61) --
	(120.55,153.05) --
	(120.62,152.50) --
	(120.68,151.94) --
	(120.75,151.38) --
	(120.81,150.83) --
	(120.88,150.27) --
	(120.94,149.72) --
	(121.00,149.16) --
	(121.07,148.61) --
	(121.13,148.05) --
	(121.19,147.50) --
	(121.25,146.94) --
	(121.30,146.38) --
	(121.36,145.83) --
	(121.42,145.27) --
	(121.47,144.72) --
	(121.52,144.16) --
	(121.57,143.61) --
	(121.62,143.05) --
	(121.66,142.49) --
	(121.71,141.94) --
	(121.75,141.38) --
	(121.79,140.83) --
	(121.83,140.27) --
	(121.86,139.72) --
	(121.89,139.16) --
	(121.92,138.61) --
	(121.95,138.05) --
	(121.97,137.49) --
	(121.99,136.94) --
	(122.00,136.38) --
	(122.02,135.83) --
	(122.03,135.27) --
	(122.03,134.72) --
	(122.03,134.16) --
	(122.03,133.60) --
	(122.02,133.05) --
	(122.01,132.49) --
	(122.00,131.94) --
	(121.98,131.38) --
	(121.96,130.83) --
	(121.93,130.27) --
	(121.90,129.72) --
	(121.87,129.16) --
	(121.84,128.60) --
	(121.80,128.05) --
	(121.75,127.49) --
	(121.71,126.94) --
	(121.66,126.38) --
	(121.61,125.83) --
	(121.55,125.27) --
	(121.50,124.71) --
	(121.44,124.16) --
	(121.38,123.60) --
	(121.32,123.05) --
	(121.26,122.49) --
	(121.20,121.94) --
	(121.14,121.38) --
	(121.09,120.82) --
	(121.03,120.27) --
	(120.97,119.71) --
	(120.91,119.16) --
	(120.86,118.60) --
	(120.81,118.05) --
	(120.76,117.49) --
	(120.71,116.94) --
	(120.66,116.38) --
	(120.62,115.82) --
	(120.58,115.27) --
	(120.54,114.71) --
	(120.50,114.16) --
	(120.46,113.60) --
	(120.42,113.05) --
	(120.39,112.49) --
	(120.36,111.93) --
	(120.32,111.38) --
	(120.29,110.82) --
	(120.26,110.27) --
	(120.23,109.71) --
	(120.19,109.16) --
	(120.16,108.60) --
	(120.12,108.05) --
	(120.09,107.49) --
	(120.05,106.93) --
	(120.01,106.38) --
	(119.97,105.82) --
	(119.92,105.27) --
	(119.88,104.71) --
	(119.83,104.16) --
	(116.59,104.16) --
	cycle;

\path[draw=drawColor,line width= 0.6pt,line join=round,line cap=round,fill=fillColor] (157.92,113.86) --
	(157.98,113.93) --
	(158.09,113.99) --
	(158.24,114.06) --
	(158.45,114.13) --
	(158.69,114.20) --
	(158.97,114.27) --
	(159.29,114.34) --
	(159.63,114.40) --
	(159.99,114.47) --
	(160.37,114.54) --
	(160.76,114.61) --
	(161.16,114.68) --
	(161.55,114.75) --
	(161.94,114.82) --
	(162.31,114.88) --
	(162.67,114.95) --
	(163.01,115.02) --
	(163.33,115.09) --
	(163.62,115.16) --
	(163.89,115.23) --
	(164.12,115.29) --
	(164.32,115.36) --
	(164.49,115.43) --
	(164.62,115.50) --
	(164.72,115.57) --
	(164.78,115.64) --
	(164.80,115.71) --
	(164.78,115.77) --
	(164.72,115.84) --
	(164.62,115.91) --
	(164.48,115.98) --
	(164.30,116.05) --
	(164.07,116.12) --
	(163.80,116.18) --
	(163.48,116.25) --
	(163.13,116.32) --
	(162.72,116.39) --
	(162.27,116.46) --
	(161.77,116.53) --
	(161.23,116.60) --
	(160.65,116.66) --
	(160.02,116.73) --
	(159.35,116.80) --
	(158.65,116.87) --
	(157.91,116.94) --
	(157.14,117.01) --
	(156.33,117.07) --
	(155.50,117.14) --
	(154.65,117.21) --
	(153.78,117.28) --
	(152.90,117.35) --
	(152.01,117.42) --
	(151.10,117.48) --
	(150.20,117.55) --
	(149.30,117.62) --
	(148.41,117.69) --
	(147.52,117.76) --
	(146.65,117.83) --
	(145.79,117.90) --
	(144.95,117.96) --
	(144.13,118.03) --
	(143.33,118.10) --
	(142.55,118.17) --
	(141.81,118.24) --
	(141.09,118.31) --
	(140.39,118.37) --
	(139.73,118.44) --
	(139.10,118.51) --
	(138.50,118.58) --
	(137.93,118.65) --
	(137.38,118.72) --
	(136.87,118.79) --
	(136.38,118.85) --
	(135.92,118.92) --
	(135.49,118.99) --
	(135.08,119.06) --
	(134.69,119.13) --
	(134.31,119.20) --
	(133.96,119.26) --
	(133.61,119.33) --
	(133.28,119.40) --
	(132.97,119.47) --
	(132.66,119.54) --
	(132.36,119.61) --
	(132.07,119.68) --
	(131.79,119.74) --
	(131.53,119.81) --
	(131.28,119.88) --
	(131.05,119.95) --
	(130.84,120.02) --
	(130.66,120.09) --
	(130.50,120.15) --
	(130.38,120.22) --
	(130.30,120.29) --
	(130.25,120.36) --
	(130.26,120.43) --
	(130.31,120.50) --
	(130.41,120.56) --
	(130.57,120.63) --
	(130.77,120.70) --
	(131.04,120.77) --
	(131.35,120.84) --
	(131.72,120.91) --
	(132.14,120.98) --
	(132.60,121.04) --
	(133.12,121.11) --
	(133.67,121.18) --
	(134.27,121.25) --
	(134.90,121.32) --
	(135.57,121.39) --
	(136.27,121.45) --
	(137.00,121.52) --
	(137.77,121.59) --
	(138.56,121.66) --
	(139.37,121.73) --
	(140.22,121.80) --
	(141.09,121.87) --
	(141.98,121.93) --
	(142.90,122.00) --
	(143.84,122.07) --
	(144.81,122.14) --
	(187.93,122.14) --
	(188.90,122.07) --
	(189.84,122.00) --
	(190.76,121.93) --
	(191.65,121.87) --
	(192.52,121.80) --
	(193.37,121.73) --
	(194.18,121.66) --
	(194.97,121.59) --
	(195.74,121.52) --
	(196.47,121.45) --
	(197.17,121.39) --
	(197.84,121.32) --
	(198.47,121.25) --
	(199.07,121.18) --
	(199.62,121.11) --
	(200.14,121.04) --
	(200.60,120.98) --
	(201.02,120.91) --
	(201.39,120.84) --
	(201.70,120.77) --
	(201.96,120.70) --
	(202.17,120.63) --
	(202.33,120.56) --
	(202.43,120.50) --
	(202.48,120.43) --
	(202.49,120.36) --
	(202.44,120.29) --
	(202.36,120.22) --
	(202.24,120.15) --
	(202.08,120.09) --
	(201.90,120.02) --
	(201.69,119.95) --
	(201.46,119.88) --
	(201.21,119.81) --
	(200.95,119.74) --
	(200.67,119.68) --
	(200.38,119.61) --
	(200.08,119.54) --
	(199.77,119.47) --
	(199.46,119.40) --
	(199.13,119.33) --
	(198.78,119.26) --
	(198.43,119.20) --
	(198.05,119.13) --
	(197.66,119.06) --
	(197.25,118.99) --
	(196.81,118.92) --
	(196.36,118.85) --
	(195.87,118.79) --
	(195.36,118.72) --
	(194.81,118.65) --
	(194.24,118.58) --
	(193.64,118.51) --
	(193.01,118.44) --
	(192.35,118.37) --
	(191.65,118.31) --
	(190.93,118.24) --
	(190.19,118.17) --
	(189.41,118.10) --
	(188.61,118.03) --
	(187.79,117.96) --
	(186.95,117.90) --
	(186.09,117.83) --
	(185.22,117.76) --
	(184.33,117.69) --
	(183.44,117.62) --
	(182.54,117.55) --
	(181.63,117.48) --
	(180.73,117.42) --
	(179.84,117.35) --
	(178.96,117.28) --
	(178.09,117.21) --
	(177.24,117.14) --
	(176.41,117.07) --
	(175.60,117.01) --
	(174.83,116.94) --
	(174.09,116.87) --
	(173.39,116.80) --
	(172.72,116.73) --
	(172.09,116.66) --
	(171.51,116.60) --
	(170.97,116.53) --
	(170.47,116.46) --
	(170.02,116.39) --
	(169.61,116.32) --
	(169.26,116.25) --
	(168.94,116.18) --
	(168.67,116.12) --
	(168.44,116.05) --
	(168.26,115.98) --
	(168.12,115.91) --
	(168.02,115.84) --
	(167.96,115.77) --
	(167.94,115.71) --
	(167.96,115.64) --
	(168.02,115.57) --
	(168.12,115.50) --
	(168.25,115.43) --
	(168.42,115.36) --
	(168.62,115.29) --
	(168.85,115.23) --
	(169.12,115.16) --
	(169.41,115.09) --
	(169.73,115.02) --
	(170.07,114.95) --
	(170.43,114.88) --
	(170.80,114.82) --
	(171.19,114.75) --
	(171.58,114.68) --
	(171.98,114.61) --
	(172.37,114.54) --
	(172.75,114.47) --
	(173.11,114.40) --
	(173.45,114.34) --
	(173.77,114.27) --
	(174.05,114.20) --
	(174.29,114.13) --
	(174.50,114.06) --
	(174.65,113.99) --
	(174.76,113.93) --
	(174.82,113.86) --
	(157.92,113.86) --
	cycle;

\path[draw=drawColor,line width= 0.6pt,line join=round,line cap=round,fill=fillColor] (208.86, 51.06) --
	(208.55, 51.43) --
	(208.25, 51.79) --
	(207.95, 52.15) --
	(207.67, 52.51) --
	(207.39, 52.88) --
	(207.13, 53.24) --
	(206.88, 53.60) --
	(206.65, 53.97) --
	(206.45, 54.33) --
	(206.26, 54.69) --
	(206.10, 55.05) --
	(205.96, 55.42) --
	(205.85, 55.78) --
	(205.78, 56.14) --
	(205.73, 56.50) --
	(205.71, 56.87) --
	(205.73, 57.23) --
	(205.77, 57.59) --
	(205.85, 57.95) --
	(205.96, 58.32) --
	(206.09, 58.68) --
	(206.26, 59.04) --
	(206.45, 59.40) --
	(206.67, 59.77) --
	(206.90, 60.13) --
	(207.16, 60.49) --
	(207.43, 60.85) --
	(207.71, 61.22) --
	(208.00, 61.58) --
	(208.30, 61.94) --
	(208.60, 62.30) --
	(208.90, 62.67) --
	(209.19, 63.03) --
	(209.48, 63.39) --
	(209.76, 63.75) --
	(210.04, 64.12) --
	(210.30, 64.48) --
	(210.54, 64.84) --
	(210.77, 65.21) --
	(210.99, 65.57) --
	(211.19, 65.93) --
	(211.37, 66.29) --
	(211.54, 66.66) --
	(211.69, 67.02) --
	(211.83, 67.38) --
	(211.96, 67.74) --
	(212.07, 68.11) --
	(212.18, 68.47) --
	(212.27, 68.83) --
	(212.36, 69.19) --
	(212.44, 69.56) --
	(212.51, 69.92) --
	(212.59, 70.28) --
	(212.65, 70.64) --
	(212.72, 71.01) --
	(212.79, 71.37) --
	(212.85, 71.73) --
	(212.92, 72.09) --
	(212.98, 72.46) --
	(213.05, 72.82) --
	(213.12, 73.18) --
	(213.19, 73.54) --
	(213.26, 73.91) --
	(213.32, 74.27) --
	(213.39, 74.63) --
	(213.46, 74.99) --
	(213.53, 75.36) --
	(213.59, 75.72) --
	(213.65, 76.08) --
	(213.71, 76.44) --
	(213.77, 76.81) --
	(213.82, 77.17) --
	(213.86, 77.53) --
	(213.90, 77.90) --
	(213.93, 78.26) --
	(213.96, 78.62) --
	(213.97, 78.98) --
	(213.99, 79.35) --
	(213.99, 79.71) --
	(213.98, 80.07) --
	(213.97, 80.43) --
	(213.95, 80.80) --
	(213.92, 81.16) --
	(213.88, 81.52) --
	(213.83, 81.88) --
	(213.77, 82.25) --
	(213.71, 82.61) --
	(213.64, 82.97) --
	(213.56, 83.33) --
	(213.47, 83.70) --
	(213.38, 84.06) --
	(213.28, 84.42) --
	(213.18, 84.78) --
	(213.07, 85.15) --
	(212.96, 85.51) --
	(212.85, 85.87) --
	(212.73, 86.23) --
	(212.62, 86.60) --
	(212.51, 86.96) --
	(212.40, 87.32) --
	(212.30, 87.68) --
	(212.20, 88.05) --
	(212.11, 88.41) --
	(212.03, 88.77) --
	(211.95, 89.14) --
	(211.89, 89.50) --
	(211.83, 89.86) --
	(211.79, 90.22) --
	(211.76, 90.59) --
	(211.74, 90.95) --
	(211.73, 91.31) --
	(211.74, 91.67) --
	(211.76, 92.04) --
	(211.80, 92.40) --
	(211.84, 92.76) --
	(217.21, 92.76) --
	(217.26, 92.40) --
	(217.29, 92.04) --
	(217.31, 91.67) --
	(217.32, 91.31) --
	(217.32, 90.95) --
	(217.30, 90.59) --
	(217.27, 90.22) --
	(217.22, 89.86) --
	(217.17, 89.50) --
	(217.10, 89.14) --
	(217.03, 88.77) --
	(216.95, 88.41) --
	(216.85, 88.05) --
	(216.76, 87.68) --
	(216.65, 87.32) --
	(216.54, 86.96) --
	(216.43, 86.60) --
	(216.32, 86.23) --
	(216.21, 85.87) --
	(216.10, 85.51) --
	(215.99, 85.15) --
	(215.88, 84.78) --
	(215.78, 84.42) --
	(215.68, 84.06) --
	(215.58, 83.70) --
	(215.50, 83.33) --
	(215.42, 82.97) --
	(215.35, 82.61) --
	(215.28, 82.25) --
	(215.23, 81.88) --
	(215.18, 81.52) --
	(215.14, 81.16) --
	(215.11, 80.80) --
	(215.09, 80.43) --
	(215.07, 80.07) --
	(215.07, 79.71) --
	(215.07, 79.35) --
	(215.08, 78.98) --
	(215.10, 78.62) --
	(215.12, 78.26) --
	(215.16, 77.90) --
	(215.20, 77.53) --
	(215.24, 77.17) --
	(215.29, 76.81) --
	(215.34, 76.44) --
	(215.40, 76.08) --
	(215.46, 75.72) --
	(215.53, 75.36) --
	(215.59, 74.99) --
	(215.66, 74.63) --
	(215.73, 74.27) --
	(215.80, 73.91) --
	(215.87, 73.54) --
	(215.94, 73.18) --
	(216.01, 72.82) --
	(216.07, 72.46) --
	(216.14, 72.09) --
	(216.20, 71.73) --
	(216.27, 71.37) --
	(216.34, 71.01) --
	(216.40, 70.64) --
	(216.47, 70.28) --
	(216.54, 69.92) --
	(216.62, 69.56) --
	(216.70, 69.19) --
	(216.78, 68.83) --
	(216.88, 68.47) --
	(216.98, 68.11) --
	(217.10, 67.74) --
	(217.22, 67.38) --
	(217.36, 67.02) --
	(217.51, 66.66) --
	(217.68, 66.29) --
	(217.87, 65.93) --
	(218.07, 65.57) --
	(218.28, 65.21) --
	(218.51, 64.84) --
	(218.76, 64.48) --
	(219.02, 64.12) --
	(219.29, 63.75) --
	(219.57, 63.39) --
	(219.86, 63.03) --
	(220.16, 62.67) --
	(220.46, 62.30) --
	(220.76, 61.94) --
	(221.06, 61.58) --
	(221.35, 61.22) --
	(221.63, 60.85) --
	(221.90, 60.49) --
	(222.15, 60.13) --
	(222.39, 59.77) --
	(222.60, 59.40) --
	(222.80, 59.04) --
	(222.96, 58.68) --
	(223.10, 58.32) --
	(223.21, 57.95) --
	(223.28, 57.59) --
	(223.33, 57.23) --
	(223.35, 56.87) --
	(223.33, 56.50) --
	(223.28, 56.14) --
	(223.20, 55.78) --
	(223.09, 55.42) --
	(222.96, 55.05) --
	(222.80, 54.69) --
	(222.61, 54.33) --
	(222.40, 53.97) --
	(222.17, 53.60) --
	(221.93, 53.24) --
	(221.66, 52.88) --
	(221.39, 52.51) --
	(221.10, 52.15) --
	(220.81, 51.79) --
	(220.50, 51.43) --
	(220.20, 51.06) --
	(208.86, 51.06) --
	cycle;
\definecolor[named]{fillColor}{rgb}{0.00,0.00,0.00}

\path[fill=fillColor] ( 70.05,141.20) circle (  1.60);
\definecolor[named]{fillColor}{rgb}{0.20,0.20,0.20}

\path[draw=drawColor,line width= 0.6pt,line join=round,fill=fillColor] ( 70.05,167.99) -- ( 70.05,172.30);

\path[draw=drawColor,line width= 0.6pt,line join=round,fill=fillColor] ( 70.05,158.98) -- ( 70.05,145.95);
\definecolor[named]{fillColor}{rgb}{0.00,0.00,0.00}

\path[draw=drawColor,line width= 0.6pt,line join=round,line cap=round,fill=fillColor] ( 69.15,167.99) --
	( 69.15,158.98) --
	( 70.96,158.98) --
	( 70.96,167.99) --
	( 69.15,167.99) --
	cycle;
\definecolor[named]{fillColor}{rgb}{0.20,0.20,0.20}

\path[draw=drawColor,line width= 1.1pt,line join=round,fill=fillColor] ( 69.15,164.45) -- ( 70.96,164.45);

\path[draw=drawColor,line width= 0.6pt,line join=round,fill=fillColor] (118.21,144.72) -- (118.21,160.57);

\path[draw=drawColor,line width= 0.6pt,line join=round,fill=fillColor] (118.21,123.53) -- (118.21,103.86);
\definecolor[named]{fillColor}{rgb}{0.00,0.00,0.00}

\path[draw=drawColor,line width= 0.6pt,line join=round,line cap=round,fill=fillColor] (117.31,144.72) --
	(117.31,123.53) --
	(119.11,123.53) --
	(119.11,144.72) --
	(117.31,144.72) --
	cycle;
\definecolor[named]{fillColor}{rgb}{0.20,0.20,0.20}

\path[draw=drawColor,line width= 1.1pt,line join=round,fill=fillColor] (117.31,134.01) -- (119.11,134.01);
\definecolor[named]{fillColor}{rgb}{0.00,0.00,0.00}

\path[fill=fillColor] (166.37,113.81) circle (  1.60);
\definecolor[named]{fillColor}{rgb}{0.20,0.20,0.20}

\path[draw=drawColor,line width= 0.6pt,line join=round,fill=fillColor] (166.37,120.68) -- (166.37,122.18);

\path[draw=drawColor,line width= 0.6pt,line join=round,fill=fillColor] (166.37,118.62) -- (166.37,117.67);
\definecolor[named]{fillColor}{rgb}{0.00,0.00,0.00}

\path[draw=drawColor,line width= 0.6pt,line join=round,line cap=round,fill=fillColor] (165.47,120.68) --
	(165.47,118.62) --
	(167.27,118.62) --
	(167.27,120.68) --
	(165.47,120.68) --
	cycle;
\definecolor[named]{fillColor}{rgb}{0.20,0.20,0.20}

\path[draw=drawColor,line width= 1.1pt,line join=round,fill=fillColor] (165.47,119.87) -- (167.27,119.87);
\definecolor[named]{fillColor}{rgb}{0.00,0.00,0.00}

\path[fill=fillColor] (214.53, 89.54) circle (  1.60);

\path[fill=fillColor] (214.53, 93.02) circle (  1.60);
\definecolor[named]{fillColor}{rgb}{0.20,0.20,0.20}

\path[draw=drawColor,line width= 0.6pt,line join=round,fill=fillColor] (214.53, 67.94) -- (214.53, 70.60);

\path[draw=drawColor,line width= 0.6pt,line join=round,fill=fillColor] (214.53, 56.16) -- (214.53, 50.81);
\definecolor[named]{fillColor}{rgb}{0.00,0.00,0.00}

\path[draw=drawColor,line width= 0.6pt,line join=round,line cap=round,fill=fillColor] (213.62, 67.94) --
	(213.62, 56.16) --
	(215.43, 56.16) --
	(215.43, 67.94) --
	(213.62, 67.94) --
	cycle;
\definecolor[named]{fillColor}{rgb}{0.20,0.20,0.20}

\path[draw=drawColor,line width= 1.1pt,line join=round,fill=fillColor] (213.62, 58.74) -- (215.43, 58.74);
\definecolor[named]{drawColor}{rgb}{0.00,0.00,0.00}
\definecolor[named]{fillColor}{rgb}{1.00,1.00,1.00}

\path[draw=drawColor,line width= 0.4pt,line join=round,line cap=round,fill=fillColor] ( 70.05,164.45) circle (  2.13);

\path[draw=drawColor,line width= 0.4pt,line join=round,line cap=round,fill=fillColor] (118.21,134.01) circle (  2.13);

\path[draw=drawColor,line width= 0.4pt,line join=round,line cap=round,fill=fillColor] (166.37,119.87) circle (  2.13);

\path[draw=drawColor,line width= 0.4pt,line join=round,line cap=round,fill=fillColor] (214.53, 58.74) circle (  2.13);
\end{scope}
\begin{scope}
\path[clip] (  0.00,  0.00) rectangle (251.50,179.23);
\definecolor[named]{drawColor}{rgb}{0.00,0.00,0.00}

\path[draw=drawColor,line width= 0.6pt,line join=round] ( 33.94, 44.73) --
	( 33.94,178.38);
\end{scope}
\begin{scope}
\path[clip] (  0.00,  0.00) rectangle (251.50,179.23);
\definecolor[named]{drawColor}{rgb}{0.00,0.00,0.00}

\node[text=drawColor,anchor=base east,inner sep=0pt, outer sep=0pt, scale=  1.00] at ( 26.82, 63.41) {$0.6$};

\node[text=drawColor,anchor=base east,inner sep=0pt, outer sep=0pt, scale=  1.00] at ( 26.82,102.00) {$0.7$};

\node[text=drawColor,anchor=base east,inner sep=0pt, outer sep=0pt, scale=  1.00] at ( 26.82,140.58) {$0.8$};
\end{scope}
\begin{scope}
\path[clip] (  0.00,  0.00) rectangle (251.50,179.23);
\definecolor[named]{drawColor}{rgb}{0.00,0.00,0.00}

\path[draw=drawColor,line width= 0.6pt,line join=round] ( 29.67, 66.86) --
	( 33.94, 66.86);

\path[draw=drawColor,line width= 0.6pt,line join=round] ( 29.67,105.44) --
	( 33.94,105.44);

\path[draw=drawColor,line width= 0.6pt,line join=round] ( 29.67,144.02) --
	( 33.94,144.02);
\end{scope}
\begin{scope}
\path[clip] (  0.00,  0.00) rectangle (251.50,179.23);
\definecolor[named]{drawColor}{rgb}{0.00,0.00,0.00}

\path[draw=drawColor,line width= 0.6pt,line join=round] ( 33.94, 44.73) --
	(250.65, 44.73);
\end{scope}
\begin{scope}
\path[clip] (  0.00,  0.00) rectangle (251.50,179.23);
\definecolor[named]{drawColor}{rgb}{0.00,0.00,0.00}

\path[draw=drawColor,line width= 0.6pt,line join=round] ( 70.05, 40.47) --
	( 70.05, 44.73);

\path[draw=drawColor,line width= 0.6pt,line join=round] (118.21, 40.47) --
	(118.21, 44.73);

\path[draw=drawColor,line width= 0.6pt,line join=round] (166.37, 40.47) --
	(166.37, 44.73);

\path[draw=drawColor,line width= 0.6pt,line join=round] (214.53, 40.47) --
	(214.53, 44.73);
\end{scope}
\begin{scope}
\path[clip] (  0.00,  0.00) rectangle (251.50,179.23);
\definecolor[named]{drawColor}{rgb}{0.00,0.00,0.00}

\node[text=drawColor,rotate= 30.00,anchor=base east,inner sep=0pt, outer sep=0pt, scale=  1.00] at ( 73.50, 31.66) {rand-8x4-inv};

\node[text=drawColor,rotate= 30.00,anchor=base east,inner sep=0pt, outer sep=0pt, scale=  1.00] at (121.66, 31.66) {rand-8x4-neg};

\node[text=drawColor,rotate= 30.00,anchor=base east,inner sep=0pt, outer sep=0pt, scale=  1.00] at (169.81, 31.66) {all-1-inv};

\node[text=drawColor,rotate= 30.00,anchor=base east,inner sep=0pt, outer sep=0pt, scale=  1.00] at (217.97, 31.66) {all-1-neg};
\end{scope}
\begin{scope}
\path[clip] (  0.00,  0.00) rectangle (251.50,179.23);
\definecolor[named]{drawColor}{rgb}{0.00,0.00,0.00}

\node[text=drawColor,rotate= 90.00,anchor=base,inner sep=0pt, outer sep=0pt, scale=  1.00] at (  7.08,111.56) {performance};
\end{scope}
\end{tikzpicture}

%% file: img/all-vs-rand.tex
\begin{tikzpicture}[x=1pt,y=1pt]
\definecolor[named]{fillColor}{rgb}{1.00,1.00,1.00}
\path[use as bounding box,fill=fillColor,fill opacity=0.00] (0,0) rectangle (251.50,179.23);
\begin{scope}
\path[clip] (  0.00,  0.00) rectangle (251.50,179.23);
\definecolor[named]{drawColor}{rgb}{1.00,1.00,1.00}
\definecolor[named]{fillColor}{rgb}{1.00,1.00,1.00}

\path[draw=drawColor,line width= 0.6pt,line join=round,line cap=round,fill=fillColor] (  0.00,  0.00) rectangle (251.50,179.23);
\end{scope}
\begin{scope}
\path[clip] ( 38.93, 46.40) rectangle (250.65,178.38);
\definecolor[named]{drawColor}{rgb}{0.20,0.20,0.20}
\definecolor[named]{fillColor}{rgb}{0.75,0.75,0.75}

\path[draw=drawColor,line width= 0.6pt,line join=round,line cap=round,fill=fillColor] ( 57.31,153.90) --
	( 57.07,154.10) --
	( 56.82,154.31) --
	( 56.58,154.51) --
	( 56.34,154.72) --
	( 56.10,154.92) --
	( 55.86,155.13) --
	( 55.63,155.34) --
	( 55.40,155.54) --
	( 55.18,155.75) --
	( 54.97,155.95) --
	( 54.76,156.16) --
	( 54.57,156.36) --
	( 54.38,156.57) --
	( 54.20,156.77) --
	( 54.03,156.98) --
	( 53.88,157.18) --
	( 53.74,157.39) --
	( 53.61,157.60) --
	( 53.50,157.80) --
	( 53.40,158.01) --
	( 53.32,158.21) --
	( 53.26,158.42) --
	( 53.21,158.62) --
	( 53.18,158.83) --
	( 53.16,159.03) --
	( 53.17,159.24) --
	( 53.19,159.44) --
	( 53.22,159.65) --
	( 53.28,159.86) --
	( 53.35,160.06) --
	( 53.43,160.27) --
	( 53.53,160.47) --
	( 53.64,160.68) --
	( 53.76,160.88) --
	( 53.89,161.09) --
	( 54.04,161.29) --
	( 54.18,161.50) --
	( 54.34,161.70) --
	( 54.50,161.91) --
	( 54.66,162.12) --
	( 54.82,162.32) --
	( 54.98,162.53) --
	( 55.14,162.73) --
	( 55.29,162.94) --
	( 55.44,163.14) --
	( 55.58,163.35) --
	( 55.71,163.55) --
	( 55.83,163.76) --
	( 55.94,163.96) --
	( 56.03,164.17) --
	( 56.12,164.38) --
	( 56.19,164.58) --
	( 56.24,164.79) --
	( 56.28,164.99) --
	( 56.31,165.20) --
	( 56.33,165.40) --
	( 56.33,165.61) --
	( 56.31,165.81) --
	( 56.29,166.02) --
	( 56.26,166.22) --
	( 56.21,166.43) --
	( 56.16,166.64) --
	( 56.10,166.84) --
	( 56.03,167.05) --
	( 55.96,167.25) --
	( 55.89,167.46) --
	( 55.82,167.66) --
	( 55.75,167.87) --
	( 55.68,168.07) --
	( 55.62,168.28) --
	( 55.56,168.49) --
	( 55.51,168.69) --
	( 55.47,168.90) --
	( 55.44,169.10) --
	( 55.42,169.31) --
	( 55.42,169.51) --
	( 55.43,169.72) --
	( 55.45,169.92) --
	( 55.49,170.13) --
	( 55.54,170.33) --
	( 55.61,170.54) --
	( 55.69,170.75) --
	( 55.79,170.95) --
	( 55.91,171.16) --
	( 56.04,171.36) --
	( 56.18,171.57) --
	( 56.34,171.77) --
	( 56.51,171.98) --
	( 56.69,172.18) --
	( 70.04,172.18) --
	( 70.22,171.98) --
	( 70.39,171.77) --
	( 70.54,171.57) --
	( 70.69,171.36) --
	( 70.82,171.16) --
	( 70.93,170.95) --
	( 71.03,170.75) --
	( 71.12,170.54) --
	( 71.18,170.33) --
	( 71.24,170.13) --
	( 71.27,169.92) --
	( 71.30,169.72) --
	( 71.31,169.51) --
	( 71.30,169.31) --
	( 71.28,169.10) --
	( 71.25,168.90) --
	( 71.21,168.69) --
	( 71.16,168.49) --
	( 71.11,168.28) --
	( 71.04,168.07) --
	( 70.98,167.87) --
	( 70.91,167.66) --
	( 70.83,167.46) --
	( 70.76,167.25) --
	( 70.69,167.05) --
	( 70.63,166.84) --
	( 70.57,166.64) --
	( 70.51,166.43) --
	( 70.47,166.22) --
	( 70.43,166.02) --
	( 70.41,165.81) --
	( 70.40,165.61) --
	( 70.40,165.40) --
	( 70.41,165.20) --
	( 70.44,164.99) --
	( 70.48,164.79) --
	( 70.54,164.58) --
	( 70.61,164.38) --
	( 70.69,164.17) --
	( 70.79,163.96) --
	( 70.90,163.76) --
	( 71.02,163.55) --
	( 71.15,163.35) --
	( 71.29,163.14) --
	( 71.43,162.94) --
	( 71.59,162.73) --
	( 71.74,162.53) --
	( 71.90,162.32) --
	( 72.07,162.12) --
	( 72.23,161.91) --
	( 72.39,161.70) --
	( 72.54,161.50) --
	( 72.69,161.29) --
	( 72.83,161.09) --
	( 72.96,160.88) --
	( 73.09,160.68) --
	( 73.20,160.47) --
	( 73.30,160.27) --
	( 73.38,160.06) --
	( 73.45,159.86) --
	( 73.50,159.65) --
	( 73.54,159.44) --
	( 73.56,159.24) --
	( 73.56,159.03) --
	( 73.55,158.83) --
	( 73.52,158.62) --
	( 73.47,158.42) --
	( 73.40,158.21) --
	( 73.32,158.01) --
	( 73.23,157.80) --
	( 73.11,157.60) --
	( 72.99,157.39) --
	( 72.85,157.18) --
	( 72.69,156.98) --
	( 72.53,156.77) --
	( 72.35,156.57) --
	( 72.16,156.36) --
	( 71.96,156.16) --
	( 71.76,155.95) --
	( 71.54,155.75) --
	( 71.32,155.54) --
	( 71.10,155.34) --
	( 70.86,155.13) --
	( 70.63,154.92) --
	( 70.39,154.72) --
	( 70.15,154.51) --
	( 69.90,154.31) --
	( 69.66,154.10) --
	( 69.42,153.90) --
	( 57.31,153.90) --
	cycle;

\path[draw=drawColor,line width= 0.6pt,line join=round,line cap=round,fill=fillColor] ( 94.10, 95.85) --
	( 94.07, 96.44) --
	( 94.04, 97.02) --
	( 94.02, 97.61) --
	( 94.00, 98.20) --
	( 93.99, 98.78) --
	( 93.98, 99.37) --
	( 93.97, 99.96) --
	( 93.96,100.54) --
	( 93.96,101.13) --
	( 93.96,101.72) --
	( 93.96,102.30) --
	( 93.96,102.89) --
	( 93.96,103.47) --
	( 93.96,104.06) --
	( 93.96,104.65) --
	( 93.97,105.23) --
	( 93.97,105.82) --
	( 93.97,106.41) --
	( 93.97,106.99) --
	( 93.97,107.58) --
	( 93.96,108.17) --
	( 93.96,108.75) --
	( 93.95,109.34) --
	( 93.94,109.93) --
	( 93.92,110.51) --
	( 93.90,111.10) --
	( 93.88,111.69) --
	( 93.86,112.27) --
	( 93.83,112.86) --
	( 93.79,113.45) --
	( 93.76,114.03) --
	( 93.72,114.62) --
	( 93.67,115.20) --
	( 93.62,115.79) --
	( 93.57,116.38) --
	( 93.51,116.96) --
	( 93.45,117.55) --
	( 93.39,118.14) --
	( 93.32,118.72) --
	( 93.25,119.31) --
	( 93.18,119.90) --
	( 93.11,120.48) --
	( 93.03,121.07) --
	( 92.95,121.66) --
	( 92.87,122.24) --
	( 92.80,122.83) --
	( 92.72,123.42) --
	( 92.64,124.00) --
	( 92.56,124.59) --
	( 92.48,125.18) --
	( 92.41,125.76) --
	( 92.34,126.35) --
	( 92.27,126.93) --
	( 92.20,127.52) --
	( 92.14,128.11) --
	( 92.09,128.69) --
	( 92.03,129.28) --
	( 91.99,129.87) --
	( 91.95,130.45) --
	( 91.91,131.04) --
	( 91.89,131.63) --
	( 91.87,132.21) --
	( 91.85,132.80) --
	( 91.84,133.39) --
	( 91.84,133.97) --
	( 91.85,134.56) --
	( 91.87,135.15) --
	( 91.89,135.73) --
	( 91.92,136.32) --
	( 91.96,136.91) --
	( 92.00,137.49) --
	( 92.05,138.08) --
	( 92.11,138.66) --
	( 92.17,139.25) --
	( 92.24,139.84) --
	( 92.31,140.42) --
	( 92.39,141.01) --
	( 92.47,141.60) --
	( 92.56,142.18) --
	( 92.65,142.77) --
	( 92.74,143.36) --
	( 92.84,143.94) --
	( 92.93,144.53) --
	( 93.03,145.12) --
	( 93.13,145.70) --
	( 93.23,146.29) --
	( 93.33,146.88) --
	( 93.43,147.46) --
	( 93.53,148.05) --
	( 93.63,148.64) --
	( 93.73,149.22) --
	( 93.82,149.81) --
	( 93.92,150.39) --
	( 94.01,150.98) --
	( 94.10,151.57) --
	( 97.76,151.57) --
	( 97.86,150.98) --
	( 97.95,150.39) --
	( 98.04,149.81) --
	( 98.14,149.22) --
	( 98.24,148.64) --
	( 98.34,148.05) --
	( 98.44,147.46) --
	( 98.54,146.88) --
	( 98.64,146.29) --
	( 98.74,145.70) --
	( 98.84,145.12) --
	( 98.93,144.53) --
	( 99.03,143.94) --
	( 99.13,143.36) --
	( 99.22,142.77) --
	( 99.31,142.18) --
	( 99.40,141.60) --
	( 99.48,141.01) --
	( 99.56,140.42) --
	( 99.63,139.84) --
	( 99.70,139.25) --
	( 99.76,138.66) --
	( 99.82,138.08) --
	( 99.87,137.49) --
	( 99.91,136.91) --
	( 99.95,136.32) --
	( 99.98,135.73) --
	(100.00,135.15) --
	(100.02,134.56) --
	(100.02,133.97) --
	(100.02,133.39) --
	(100.02,132.80) --
	(100.00,132.21) --
	( 99.98,131.63) --
	( 99.95,131.04) --
	( 99.92,130.45) --
	( 99.88,129.87) --
	( 99.83,129.28) --
	( 99.78,128.69) --
	( 99.72,128.11) --
	( 99.66,127.52) --
	( 99.60,126.93) --
	( 99.53,126.35) --
	( 99.46,125.76) --
	( 99.38,125.18) --
	( 99.31,124.59) --
	( 99.23,124.00) --
	( 99.15,123.42) --
	( 99.07,122.83) --
	( 98.99,122.24) --
	( 98.91,121.66) --
	( 98.84,121.07) --
	( 98.76,120.48) --
	( 98.69,119.90) --
	( 98.61,119.31) --
	( 98.55,118.72) --
	( 98.48,118.14) --
	( 98.42,117.55) --
	( 98.35,116.96) --
	( 98.30,116.38) --
	( 98.25,115.79) --
	( 98.20,115.20) --
	( 98.15,114.62) --
	( 98.11,114.03) --
	( 98.07,113.45) --
	( 98.04,112.86) --
	( 98.01,112.27) --
	( 97.98,111.69) --
	( 97.96,111.10) --
	( 97.95,110.51) --
	( 97.93,109.93) --
	( 97.92,109.34) --
	( 97.91,108.75) --
	( 97.90,108.17) --
	( 97.90,107.58) --
	( 97.90,106.99) --
	( 97.90,106.41) --
	( 97.90,105.82) --
	( 97.90,105.23) --
	( 97.90,104.65) --
	( 97.91,104.06) --
	( 97.91,103.47) --
	( 97.91,102.89) --
	( 97.91,102.30) --
	( 97.91,101.72) --
	( 97.91,101.13) --
	( 97.91,100.54) --
	( 97.90, 99.96) --
	( 97.89, 99.37) --
	( 97.88, 98.78) --
	( 97.86, 98.20) --
	( 97.85, 97.61) --
	( 97.83, 97.02) --
	( 97.80, 96.44) --
	( 97.77, 95.85) --
	( 94.10, 95.85) --
	cycle;

\path[draw=drawColor,line width= 0.6pt,line join=round,line cap=round,fill=fillColor] (119.01,156.73) --
	(118.65,156.83) --
	(118.30,156.92) --
	(117.97,157.02) --
	(117.64,157.11) --
	(117.32,157.21) --
	(117.02,157.31) --
	(116.73,157.40) --
	(116.46,157.50) --
	(116.20,157.59) --
	(115.96,157.69) --
	(115.74,157.78) --
	(115.53,157.88) --
	(115.34,157.97) --
	(115.17,158.07) --
	(115.01,158.16) --
	(114.87,158.26) --
	(114.75,158.36) --
	(114.64,158.45) --
	(114.54,158.55) --
	(114.45,158.64) --
	(114.36,158.74) --
	(114.29,158.83) --
	(114.21,158.93) --
	(114.14,159.02) --
	(114.06,159.12) --
	(113.98,159.21) --
	(113.89,159.31) --
	(113.78,159.41) --
	(113.66,159.50) --
	(113.52,159.60) --
	(113.36,159.69) --
	(113.18,159.79) --
	(112.97,159.88) --
	(112.74,159.98) --
	(112.47,160.07) --
	(112.18,160.17) --
	(111.86,160.26) --
	(111.51,160.36) --
	(111.13,160.46) --
	(110.73,160.55) --
	(110.31,160.65) --
	(109.87,160.74) --
	(109.41,160.84) --
	(108.94,160.93) --
	(108.47,161.03) --
	(107.99,161.12) --
	(107.52,161.22) --
	(107.05,161.31) --
	(106.61,161.41) --
	(106.18,161.51) --
	(105.78,161.60) --
	(105.41,161.70) --
	(105.08,161.79) --
	(104.78,161.89) --
	(104.54,161.98) --
	(104.34,162.08) --
	(104.20,162.17) --
	(104.11,162.27) --
	(104.08,162.36) --
	(104.10,162.46) --
	(104.18,162.56) --
	(104.32,162.65) --
	(104.51,162.75) --
	(104.76,162.84) --
	(105.06,162.94) --
	(105.40,163.03) --
	(105.79,163.13) --
	(106.21,163.22) --
	(106.66,163.32) --
	(107.15,163.41) --
	(107.65,163.51) --
	(108.17,163.61) --
	(108.70,163.70) --
	(109.24,163.80) --
	(109.78,163.89) --
	(110.31,163.99) --
	(110.83,164.08) --
	(111.34,164.18) --
	(111.83,164.27) --
	(112.30,164.37) --
	(112.74,164.46) --
	(113.16,164.56) --
	(113.56,164.66) --
	(113.92,164.75) --
	(114.26,164.85) --
	(114.58,164.94) --
	(114.87,165.04) --
	(115.13,165.13) --
	(115.38,165.23) --
	(115.60,165.32) --
	(115.81,165.42) --
	(116.01,165.51) --
	(116.19,165.61) --
	(116.37,165.71) --
	(116.55,165.80) --
	(116.72,165.90) --
	(116.90,165.99) --
	(117.08,166.09) --
	(117.28,166.18) --
	(117.48,166.28) --
	(117.69,166.37) --
	(117.92,166.47) --
	(118.16,166.56) --
	(118.42,166.66) --
	(118.69,166.76) --
	(138.32,166.76) --
	(138.59,166.66) --
	(138.85,166.56) --
	(139.09,166.47) --
	(139.32,166.37) --
	(139.53,166.28) --
	(139.73,166.18) --
	(139.93,166.09) --
	(140.11,165.99) --
	(140.29,165.90) --
	(140.46,165.80) --
	(140.64,165.71) --
	(140.82,165.61) --
	(141.00,165.51) --
	(141.20,165.42) --
	(141.41,165.32) --
	(141.63,165.23) --
	(141.88,165.13) --
	(142.14,165.04) --
	(142.43,164.94) --
	(142.75,164.85) --
	(143.09,164.75) --
	(143.45,164.66) --
	(143.85,164.56) --
	(144.27,164.46) --
	(144.71,164.37) --
	(145.18,164.27) --
	(145.67,164.18) --
	(146.18,164.08) --
	(146.70,163.99) --
	(147.23,163.89) --
	(147.77,163.80) --
	(148.30,163.70) --
	(148.84,163.61) --
	(149.36,163.51) --
	(149.86,163.41) --
	(150.34,163.32) --
	(150.80,163.22) --
	(151.22,163.13) --
	(151.61,163.03) --
	(151.95,162.94) --
	(152.25,162.84) --
	(152.50,162.75) --
	(152.69,162.65) --
	(152.83,162.56) --
	(152.91,162.46) --
	(152.93,162.36) --
	(152.90,162.27) --
	(152.81,162.17) --
	(152.67,162.08) --
	(152.47,161.98) --
	(152.23,161.89) --
	(151.93,161.79) --
	(151.60,161.70) --
	(151.23,161.60) --
	(150.83,161.51) --
	(150.40,161.41) --
	(149.95,161.31) --
	(149.49,161.22) --
	(149.02,161.12) --
	(148.54,161.03) --
	(148.07,160.93) --
	(147.60,160.84) --
	(147.14,160.74) --
	(146.70,160.65) --
	(146.28,160.55) --
	(145.88,160.46) --
	(145.50,160.36) --
	(145.15,160.26) --
	(144.83,160.17) --
	(144.54,160.07) --
	(144.27,159.98) --
	(144.04,159.88) --
	(143.83,159.79) --
	(143.65,159.69) --
	(143.49,159.60) --
	(143.35,159.50) --
	(143.23,159.41) --
	(143.12,159.31) --
	(143.03,159.21) --
	(142.95,159.12) --
	(142.87,159.02) --
	(142.80,158.93) --
	(142.72,158.83) --
	(142.65,158.74) --
	(142.56,158.64) --
	(142.47,158.55) --
	(142.37,158.45) --
	(142.26,158.36) --
	(142.14,158.26) --
	(142.00,158.16) --
	(141.84,158.07) --
	(141.67,157.97) --
	(141.48,157.88) --
	(141.27,157.78) --
	(141.05,157.69) --
	(140.81,157.59) --
	(140.55,157.50) --
	(140.28,157.40) --
	(139.99,157.31) --
	(139.69,157.21) --
	(139.37,157.11) --
	(139.04,157.02) --
	(138.71,156.92) --
	(138.36,156.83) --
	(138.00,156.73) --
	(119.01,156.73) --
	cycle;

\path[draw=drawColor,line width= 0.6pt,line join=round,line cap=round,fill=fillColor] (158.88, 85.33) --
	(158.83, 85.71) --
	(158.78, 86.08) --
	(158.75, 86.45) --
	(158.71, 86.82) --
	(158.69, 87.20) --
	(158.67, 87.57) --
	(158.66, 87.94) --
	(158.65, 88.31) --
	(158.66, 88.68) --
	(158.67, 89.06) --
	(158.69, 89.43) --
	(158.71, 89.80) --
	(158.75, 90.17) --
	(158.78, 90.54) --
	(158.83, 90.92) --
	(158.88, 91.29) --
	(158.94, 91.66) --
	(159.00, 92.03) --
	(159.06, 92.40) --
	(159.13, 92.78) --
	(159.21, 93.15) --
	(159.28, 93.52) --
	(159.36, 93.89) --
	(159.44, 94.26) --
	(159.51, 94.64) --
	(159.59, 95.01) --
	(159.67, 95.38) --
	(159.74, 95.75) --
	(159.81, 96.13) --
	(159.87, 96.50) --
	(159.93, 96.87) --
	(159.98, 97.24) --
	(160.03, 97.61) --
	(160.07, 97.99) --
	(160.10, 98.36) --
	(160.13, 98.73) --
	(160.14, 99.10) --
	(160.15, 99.47) --
	(160.14, 99.85) --
	(160.13,100.22) --
	(160.10,100.59) --
	(160.07,100.96) --
	(160.02,101.33) --
	(159.97,101.71) --
	(159.91,102.08) --
	(159.83,102.45) --
	(159.75,102.82) --
	(159.65,103.19) --
	(159.55,103.57) --
	(159.44,103.94) --
	(159.32,104.31) --
	(159.19,104.68) --
	(159.06,105.06) --
	(158.92,105.43) --
	(158.77,105.80) --
	(158.62,106.17) --
	(158.47,106.54) --
	(158.31,106.92) --
	(158.14,107.29) --
	(157.97,107.66) --
	(157.80,108.03) --
	(157.63,108.40) --
	(157.45,108.78) --
	(157.28,109.15) --
	(157.10,109.52) --
	(156.92,109.89) --
	(156.74,110.26) --
	(156.57,110.64) --
	(156.39,111.01) --
	(156.22,111.38) --
	(156.05,111.75) --
	(155.88,112.12) --
	(155.71,112.50) --
	(155.55,112.87) --
	(155.40,113.24) --
	(155.25,113.61) --
	(155.10,113.99) --
	(154.97,114.36) --
	(154.84,114.73) --
	(154.72,115.10) --
	(154.60,115.47) --
	(154.50,115.85) --
	(154.41,116.22) --
	(154.32,116.59) --
	(154.25,116.96) --
	(154.19,117.33) --
	(154.14,117.71) --
	(154.10,118.08) --
	(154.07,118.45) --
	(154.05,118.82) --
	(154.05,119.19) --
	(154.05,119.57) --
	(154.07,119.94) --
	(154.09,120.31) --
	(154.13,120.68) --
	(154.18,121.05) --
	(154.23,121.43) --
	(154.30,121.80) --
	(154.37,122.17) --
	(154.45,122.54) --
	(154.54,122.92) --
	(154.64,123.29) --
	(154.74,123.66) --
	(154.86,124.03) --
	(154.98,124.40) --
	(155.11,124.78) --
	(155.24,125.15) --
	(155.39,125.52) --
	(155.54,125.89) --
	(155.69,126.26) --
	(155.85,126.64) --
	(156.02,127.01) --
	(156.20,127.38) --
	(165.95,127.38) --
	(166.13,127.01) --
	(166.30,126.64) --
	(166.46,126.26) --
	(166.62,125.89) --
	(166.77,125.52) --
	(166.91,125.15) --
	(167.04,124.78) --
	(167.17,124.40) --
	(167.29,124.03) --
	(167.41,123.66) --
	(167.51,123.29) --
	(167.61,122.92) --
	(167.70,122.54) --
	(167.78,122.17) --
	(167.86,121.80) --
	(167.92,121.43) --
	(167.98,121.05) --
	(168.02,120.68) --
	(168.06,120.31) --
	(168.08,119.94) --
	(168.10,119.57) --
	(168.10,119.19) --
	(168.10,118.82) --
	(168.08,118.45) --
	(168.05,118.08) --
	(168.01,117.71) --
	(167.96,117.33) --
	(167.90,116.96) --
	(167.83,116.59) --
	(167.75,116.22) --
	(167.65,115.85) --
	(167.55,115.47) --
	(167.44,115.10) --
	(167.31,114.73) --
	(167.19,114.36) --
	(167.05,113.99) --
	(166.90,113.61) --
	(166.75,113.24) --
	(166.60,112.87) --
	(166.44,112.50) --
	(166.27,112.12) --
	(166.11,111.75) --
	(165.93,111.38) --
	(165.76,111.01) --
	(165.58,110.64) --
	(165.41,110.26) --
	(165.23,109.89) --
	(165.05,109.52) --
	(164.88,109.15) --
	(164.70,108.78) --
	(164.52,108.40) --
	(164.35,108.03) --
	(164.18,107.66) --
	(164.01,107.29) --
	(163.85,106.92) --
	(163.69,106.54) --
	(163.53,106.17) --
	(163.38,105.80) --
	(163.23,105.43) --
	(163.09,105.06) --
	(162.96,104.68) --
	(162.83,104.31) --
	(162.71,103.94) --
	(162.60,103.57) --
	(162.50,103.19) --
	(162.40,102.82) --
	(162.32,102.45) --
	(162.25,102.08) --
	(162.18,101.71) --
	(162.13,101.33) --
	(162.08,100.96) --
	(162.05,100.59) --
	(162.02,100.22) --
	(162.01, 99.85) --
	(162.01, 99.47) --
	(162.01, 99.10) --
	(162.03, 98.73) --
	(162.05, 98.36) --
	(162.08, 97.99) --
	(162.12, 97.61) --
	(162.17, 97.24) --
	(162.22, 96.87) --
	(162.28, 96.50) --
	(162.35, 96.13) --
	(162.41, 95.75) --
	(162.49, 95.38) --
	(162.56, 95.01) --
	(162.64, 94.64) --
	(162.71, 94.26) --
	(162.79, 93.89) --
	(162.87, 93.52) --
	(162.94, 93.15) --
	(163.02, 92.78) --
	(163.09, 92.40) --
	(163.15, 92.03) --
	(163.21, 91.66) --
	(163.27, 91.29) --
	(163.32, 90.92) --
	(163.37, 90.54) --
	(163.41, 90.17) --
	(163.44, 89.80) --
	(163.46, 89.43) --
	(163.48, 89.06) --
	(163.49, 88.68) --
	(163.50, 88.31) --
	(163.49, 87.94) --
	(163.48, 87.57) --
	(163.46, 87.20) --
	(163.44, 86.82) --
	(163.41, 86.45) --
	(163.37, 86.08) --
	(163.32, 85.71) --
	(163.27, 85.33) --
	(158.88, 85.33) --
	cycle;

\path[draw=drawColor,line width= 0.6pt,line join=round,line cap=round,fill=fillColor] (187.97,140.16) --
	(187.72,140.33) --
	(187.46,140.50) --
	(187.20,140.67) --
	(186.94,140.84) --
	(186.67,141.01) --
	(186.41,141.18) --
	(186.13,141.35) --
	(185.86,141.52) --
	(185.59,141.69) --
	(185.31,141.86) --
	(185.04,142.03) --
	(184.76,142.20) --
	(184.49,142.36) --
	(184.22,142.53) --
	(183.95,142.70) --
	(183.69,142.87) --
	(183.44,143.04) --
	(183.19,143.21) --
	(182.95,143.38) --
	(182.72,143.55) --
	(182.50,143.72) --
	(182.29,143.89) --
	(182.10,144.06) --
	(181.92,144.23) --
	(181.75,144.40) --
	(181.60,144.57) --
	(181.46,144.74) --
	(181.35,144.91) --
	(181.25,145.08) --
	(181.17,145.25) --
	(181.11,145.42) --
	(181.06,145.59) --
	(181.04,145.75) --
	(181.03,145.92) --
	(181.05,146.09) --
	(181.08,146.26) --
	(181.12,146.43) --
	(181.19,146.60) --
	(181.26,146.77) --
	(181.36,146.94) --
	(181.46,147.11) --
	(181.58,147.28) --
	(181.70,147.45) --
	(181.84,147.62) --
	(181.98,147.79) --
	(182.13,147.96) --
	(182.28,148.13) --
	(182.43,148.30) --
	(182.58,148.47) --
	(182.73,148.64) --
	(182.87,148.81) --
	(183.02,148.98) --
	(183.15,149.14) --
	(183.28,149.31) --
	(183.41,149.48) --
	(183.52,149.65) --
	(183.63,149.82) --
	(183.72,149.99) --
	(183.80,150.16) --
	(183.88,150.33) --
	(183.94,150.50) --
	(184.00,150.67) --
	(184.04,150.84) --
	(184.08,151.01) --
	(184.11,151.18) --
	(184.13,151.35) --
	(184.14,151.52) --
	(184.15,151.69) --
	(184.16,151.86) --
	(184.16,152.03) --
	(184.16,152.20) --
	(184.17,152.37) --
	(184.17,152.53) --
	(184.18,152.70) --
	(184.19,152.87) --
	(184.21,153.04) --
	(184.24,153.21) --
	(184.27,153.38) --
	(184.32,153.55) --
	(184.37,153.72) --
	(184.44,153.89) --
	(184.52,154.06) --
	(184.61,154.23) --
	(184.72,154.40) --
	(184.84,154.57) --
	(184.97,154.74) --
	(185.12,154.91) --
	(185.28,155.08) --
	(185.45,155.25) --
	(185.64,155.42) --
	(185.84,155.59) --
	(186.05,155.76) --
	(186.28,155.92) --
	(201.02,155.92) --
	(201.24,155.76) --
	(201.45,155.59) --
	(201.65,155.42) --
	(201.84,155.25) --
	(202.01,155.08) --
	(202.18,154.91) --
	(202.32,154.74) --
	(202.46,154.57) --
	(202.58,154.40) --
	(202.68,154.23) --
	(202.78,154.06) --
	(202.86,153.89) --
	(202.92,153.72) --
	(202.98,153.55) --
	(203.02,153.38) --
	(203.06,153.21) --
	(203.08,153.04) --
	(203.10,152.87) --
	(203.11,152.70) --
	(203.12,152.53) --
	(203.13,152.37) --
	(203.13,152.20) --
	(203.13,152.03) --
	(203.13,151.86) --
	(203.14,151.69) --
	(203.15,151.52) --
	(203.16,151.35) --
	(203.19,151.18) --
	(203.21,151.01) --
	(203.25,150.84) --
	(203.29,150.67) --
	(203.35,150.50) --
	(203.41,150.33) --
	(203.49,150.16) --
	(203.57,149.99) --
	(203.67,149.82) --
	(203.77,149.65) --
	(203.89,149.48) --
	(204.01,149.31) --
	(204.14,149.14) --
	(204.28,148.98) --
	(204.42,148.81) --
	(204.57,148.64) --
	(204.72,148.47) --
	(204.87,148.30) --
	(205.02,148.13) --
	(205.17,147.96) --
	(205.31,147.79) --
	(205.45,147.62) --
	(205.59,147.45) --
	(205.72,147.28) --
	(205.83,147.11) --
	(205.94,146.94) --
	(206.03,146.77) --
	(206.11,146.60) --
	(206.17,146.43) --
	(206.22,146.26) --
	(206.25,146.09) --
	(206.26,145.92) --
	(206.25,145.75) --
	(206.23,145.59) --
	(206.19,145.42) --
	(206.12,145.25) --
	(206.04,145.08) --
	(205.94,144.91) --
	(205.83,144.74) --
	(205.69,144.57) --
	(205.54,144.40) --
	(205.38,144.23) --
	(205.20,144.06) --
	(205.00,143.89) --
	(204.79,143.72) --
	(204.57,143.55) --
	(204.34,143.38) --
	(204.10,143.21) --
	(203.85,143.04) --
	(203.60,142.87) --
	(203.34,142.70) --
	(203.07,142.53) --
	(202.80,142.36) --
	(202.53,142.20) --
	(202.26,142.03) --
	(201.98,141.86) --
	(201.71,141.69) --
	(201.43,141.52) --
	(201.16,141.35) --
	(200.89,141.18) --
	(200.62,141.01) --
	(200.35,140.84) --
	(200.09,140.67) --
	(199.83,140.50) --
	(199.57,140.33) --
	(199.32,140.16) --
	(187.97,140.16) --
	cycle;

\path[draw=drawColor,line width= 0.6pt,line join=round,line cap=round,fill=fillColor] (224.88, 52.81) --
	(224.85, 53.54) --
	(224.81, 54.28) --
	(224.78, 55.02) --
	(224.75, 55.75) --
	(224.72, 56.49) --
	(224.70, 57.23) --
	(224.67, 57.96) --
	(224.64, 58.70) --
	(224.61, 59.44) --
	(224.58, 60.17) --
	(224.56, 60.91) --
	(224.53, 61.65) --
	(224.50, 62.38) --
	(224.47, 63.12) --
	(224.44, 63.86) --
	(224.41, 64.60) --
	(224.38, 65.33) --
	(224.35, 66.07) --
	(224.32, 66.81) --
	(224.28, 67.54) --
	(224.25, 68.28) --
	(224.21, 69.02) --
	(224.18, 69.75) --
	(224.14, 70.49) --
	(224.11, 71.23) --
	(224.07, 71.96) --
	(224.03, 72.70) --
	(224.00, 73.44) --
	(223.96, 74.17) --
	(223.93, 74.91) --
	(223.89, 75.65) --
	(223.86, 76.38) --
	(223.83, 77.12) --
	(223.80, 77.86) --
	(223.77, 78.59) --
	(223.75, 79.33) --
	(223.72, 80.07) --
	(223.70, 80.80) --
	(223.69, 81.54) --
	(223.67, 82.28) --
	(223.66, 83.02) --
	(223.65, 83.75) --
	(223.64, 84.49) --
	(223.63, 85.23) --
	(223.63, 85.96) --
	(223.63, 86.70) --
	(223.63, 87.44) --
	(223.63, 88.17) --
	(223.64, 88.91) --
	(223.65, 89.65) --
	(223.66, 90.38) --
	(223.66, 91.12) --
	(223.67, 91.86) --
	(223.69, 92.59) --
	(223.70, 93.33) --
	(223.71, 94.07) --
	(223.72, 94.80) --
	(223.73, 95.54) --
	(223.74, 96.28) --
	(223.75, 97.01) --
	(223.76, 97.75) --
	(223.77, 98.49) --
	(223.77, 99.23) --
	(223.78, 99.96) --
	(223.79,100.70) --
	(223.79,101.44) --
	(223.79,102.17) --
	(223.80,102.91) --
	(223.80,103.65) --
	(223.80,104.38) --
	(223.81,105.12) --
	(223.81,105.86) --
	(223.81,106.59) --
	(223.82,107.33) --
	(223.82,108.07) --
	(223.83,108.80) --
	(223.84,109.54) --
	(223.85,110.28) --
	(223.86,111.01) --
	(223.87,111.75) --
	(223.89,112.49) --
	(223.91,113.22) --
	(223.93,113.96) --
	(223.96,114.70) --
	(223.99,115.43) --
	(224.02,116.17) --
	(224.06,116.91) --
	(224.09,117.65) --
	(224.14,118.38) --
	(228.30,118.38) --
	(228.34,117.65) --
	(228.38,116.91) --
	(228.42,116.17) --
	(228.45,115.43) --
	(228.48,114.70) --
	(228.50,113.96) --
	(228.53,113.22) --
	(228.55,112.49) --
	(228.56,111.75) --
	(228.58,111.01) --
	(228.59,110.28) --
	(228.60,109.54) --
	(228.61,108.80) --
	(228.61,108.07) --
	(228.62,107.33) --
	(228.62,106.59) --
	(228.63,105.86) --
	(228.63,105.12) --
	(228.63,104.38) --
	(228.64,103.65) --
	(228.64,102.91) --
	(228.64,102.17) --
	(228.65,101.44) --
	(228.65,100.70) --
	(228.66, 99.96) --
	(228.66, 99.23) --
	(228.67, 98.49) --
	(228.68, 97.75) --
	(228.69, 97.01) --
	(228.70, 96.28) --
	(228.71, 95.54) --
	(228.72, 94.80) --
	(228.73, 94.07) --
	(228.74, 93.33) --
	(228.75, 92.59) --
	(228.76, 91.86) --
	(228.77, 91.12) --
	(228.78, 90.38) --
	(228.79, 89.65) --
	(228.80, 88.91) --
	(228.80, 88.17) --
	(228.80, 87.44) --
	(228.81, 86.70) --
	(228.81, 85.96) --
	(228.80, 85.23) --
	(228.80, 84.49) --
	(228.79, 83.75) --
	(228.78, 83.02) --
	(228.77, 82.28) --
	(228.75, 81.54) --
	(228.73, 80.80) --
	(228.71, 80.07) --
	(228.69, 79.33) --
	(228.66, 78.59) --
	(228.63, 77.86) --
	(228.60, 77.12) --
	(228.57, 76.38) --
	(228.54, 75.65) --
	(228.51, 74.91) --
	(228.47, 74.17) --
	(228.44, 73.44) --
	(228.40, 72.70) --
	(228.36, 71.96) --
	(228.33, 71.23) --
	(228.29, 70.49) --
	(228.26, 69.75) --
	(228.22, 69.02) --
	(228.19, 68.28) --
	(228.15, 67.54) --
	(228.12, 66.81) --
	(228.09, 66.07) --
	(228.05, 65.33) --
	(228.02, 64.60) --
	(227.99, 63.86) --
	(227.96, 63.12) --
	(227.93, 62.38) --
	(227.91, 61.65) --
	(227.88, 60.91) --
	(227.85, 60.17) --
	(227.82, 59.44) --
	(227.80, 58.70) --
	(227.77, 57.96) --
	(227.74, 57.23) --
	(227.71, 56.49) --
	(227.68, 55.75) --
	(227.65, 55.02) --
	(227.62, 54.28) --
	(227.59, 53.54) --
	(227.56, 52.81) --
	(224.88, 52.81) --
	cycle;
\definecolor[named]{fillColor}{rgb}{0.20,0.20,0.20}

\path[draw=drawColor,line width= 0.6pt,line join=round,fill=fillColor] ( 63.36,169.34) -- ( 63.36,172.38);

\path[draw=drawColor,line width= 0.6pt,line join=round,fill=fillColor] ( 63.36,159.14) -- ( 63.36,153.70);
\definecolor[named]{fillColor}{rgb}{0.00,0.00,0.00}

\path[draw=drawColor,line width= 0.6pt,line join=round,line cap=round,fill=fillColor] ( 62.75,169.34) --
	( 62.75,159.14) --
	( 63.97,159.14) --
	( 63.97,169.34) --
	( 62.75,169.34) --
	cycle;
\definecolor[named]{fillColor}{rgb}{0.20,0.20,0.20}

\path[draw=drawColor,line width= 1.1pt,line join=round,fill=fillColor] ( 62.75,160.87) -- ( 63.97,160.87);

\path[draw=drawColor,line width= 0.6pt,line join=round,fill=fillColor] ( 95.93,137.04) -- ( 95.93,151.85);

\path[draw=drawColor,line width= 0.6pt,line join=round,fill=fillColor] ( 95.93,113.27) -- ( 95.93, 95.57);
\definecolor[named]{fillColor}{rgb}{0.00,0.00,0.00}

\path[draw=drawColor,line width= 0.6pt,line join=round,line cap=round,fill=fillColor] ( 95.32,137.04) --
	( 95.32,113.27) --
	( 96.54,113.27) --
	( 96.54,137.04) --
	( 95.32,137.04) --
	cycle;
\definecolor[named]{fillColor}{rgb}{0.20,0.20,0.20}

\path[draw=drawColor,line width= 1.1pt,line join=round,fill=fillColor] ( 95.32,129.63) -- ( 96.54,129.63);

\path[draw=drawColor,line width= 0.6pt,line join=round,fill=fillColor] (128.50,163.15) -- (128.50,166.77);

\path[draw=drawColor,line width= 0.6pt,line join=round,fill=fillColor] (128.50,159.63) -- (128.50,156.72);
\definecolor[named]{fillColor}{rgb}{0.00,0.00,0.00}

\path[draw=drawColor,line width= 0.6pt,line join=round,line cap=round,fill=fillColor] (127.89,163.15) --
	(127.89,159.63) --
	(129.12,159.63) --
	(129.12,163.15) --
	(127.89,163.15) --
	cycle;
\definecolor[named]{fillColor}{rgb}{0.20,0.20,0.20}

\path[draw=drawColor,line width= 1.1pt,line join=round,fill=fillColor] (127.89,162.14) -- (129.12,162.14);
\definecolor[named]{fillColor}{rgb}{0.00,0.00,0.00}

\path[fill=fillColor] (161.08, 85.04) circle (  1.60);
\definecolor[named]{fillColor}{rgb}{0.20,0.20,0.20}

\path[draw=drawColor,line width= 0.6pt,line join=round,fill=fillColor] (161.08,121.76) -- (161.08,127.68);

\path[draw=drawColor,line width= 0.6pt,line join=round,fill=fillColor] (161.08,109.41) -- (161.08, 91.55);
\definecolor[named]{fillColor}{rgb}{0.00,0.00,0.00}

\path[draw=drawColor,line width= 0.6pt,line join=round,line cap=round,fill=fillColor] (160.47,121.76) --
	(160.47,109.41) --
	(161.69,109.41) --
	(161.69,121.76) --
	(160.47,121.76) --
	cycle;
\definecolor[named]{fillColor}{rgb}{0.20,0.20,0.20}

\path[draw=drawColor,line width= 1.1pt,line join=round,fill=fillColor] (160.47,116.91) -- (161.69,116.91);

\path[draw=drawColor,line width= 0.6pt,line join=round,fill=fillColor] (193.65,153.07) -- (193.65,156.03);

\path[draw=drawColor,line width= 0.6pt,line join=round,fill=fillColor] (193.65,144.83) -- (193.65,140.06);
\definecolor[named]{fillColor}{rgb}{0.00,0.00,0.00}

\path[draw=drawColor,line width= 0.6pt,line join=round,line cap=round,fill=fillColor] (193.04,153.07) --
	(193.04,144.83) --
	(194.26,144.83) --
	(194.26,153.07) --
	(193.04,153.07) --
	cycle;
\definecolor[named]{fillColor}{rgb}{0.20,0.20,0.20}

\path[draw=drawColor,line width= 1.1pt,line join=round,fill=fillColor] (193.04,147.14) -- (194.26,147.14);

\path[draw=drawColor,line width= 0.6pt,line join=round,fill=fillColor] (226.22,110.13) -- (226.22,118.79);

\path[draw=drawColor,line width= 0.6pt,line join=round,fill=fillColor] (226.22, 78.57) -- (226.22, 52.40);
\definecolor[named]{fillColor}{rgb}{0.00,0.00,0.00}

\path[draw=drawColor,line width= 0.6pt,line join=round,line cap=round,fill=fillColor] (225.61,110.13) --
	(225.61, 78.57) --
	(226.83, 78.57) --
	(226.83,110.13) --
	(225.61,110.13) --
	cycle;
\definecolor[named]{fillColor}{rgb}{0.20,0.20,0.20}

\path[draw=drawColor,line width= 1.1pt,line join=round,fill=fillColor] (225.61, 87.53) -- (226.83, 87.53);
\definecolor[named]{drawColor}{rgb}{0.00,0.00,0.00}
\definecolor[named]{fillColor}{rgb}{1.00,1.00,1.00}

\path[draw=drawColor,line width= 0.4pt,line join=round,line cap=round,fill=fillColor] ( 63.36,160.87) circle (  2.13);

\path[draw=drawColor,line width= 0.4pt,line join=round,line cap=round,fill=fillColor] ( 95.93,129.63) circle (  2.13);

\path[draw=drawColor,line width= 0.4pt,line join=round,line cap=round,fill=fillColor] (128.50,162.14) circle (  2.13);

\path[draw=drawColor,line width= 0.4pt,line join=round,line cap=round,fill=fillColor] (161.08,116.91) circle (  2.13);

\path[draw=drawColor,line width= 0.4pt,line join=round,line cap=round,fill=fillColor] (193.65,147.14) circle (  2.13);

\path[draw=drawColor,line width= 0.4pt,line join=round,line cap=round,fill=fillColor] (226.22, 87.53) circle (  2.13);
\end{scope}
\begin{scope}
\path[clip] (  0.00,  0.00) rectangle (251.50,179.23);
\definecolor[named]{drawColor}{rgb}{0.00,0.00,0.00}

\path[draw=drawColor,line width= 0.6pt,line join=round] ( 38.93, 46.40) --
	( 38.93,178.38);
\end{scope}
\begin{scope}
\path[clip] (  0.00,  0.00) rectangle (251.50,179.23);
\definecolor[named]{drawColor}{rgb}{0.00,0.00,0.00}

\node[text=drawColor,anchor=base east,inner sep=0pt, outer sep=0pt, scale=  1.00] at ( 31.82, 59.01) {$0.75$};

\node[text=drawColor,anchor=base east,inner sep=0pt, outer sep=0pt, scale=  1.00] at ( 31.82, 85.46) {$0.80$};

\node[text=drawColor,anchor=base east,inner sep=0pt, outer sep=0pt, scale=  1.00] at ( 31.82,111.91) {$0.85$};

\node[text=drawColor,anchor=base east,inner sep=0pt, outer sep=0pt, scale=  1.00] at ( 31.82,138.36) {$0.90$};

\node[text=drawColor,anchor=base east,inner sep=0pt, outer sep=0pt, scale=  1.00] at ( 31.82,164.81) {$0.95$};
\end{scope}
\begin{scope}
\path[clip] (  0.00,  0.00) rectangle (251.50,179.23);
\definecolor[named]{drawColor}{rgb}{0.00,0.00,0.00}

\path[draw=drawColor,line width= 0.6pt,line join=round] ( 34.67, 62.45) --
	( 38.93, 62.45);

\path[draw=drawColor,line width= 0.6pt,line join=round] ( 34.67, 88.90) --
	( 38.93, 88.90);

\path[draw=drawColor,line width= 0.6pt,line join=round] ( 34.67,115.35) --
	( 38.93,115.35);

\path[draw=drawColor,line width= 0.6pt,line join=round] ( 34.67,141.80) --
	( 38.93,141.80);

\path[draw=drawColor,line width= 0.6pt,line join=round] ( 34.67,168.25) --
	( 38.93,168.25);
\end{scope}
\begin{scope}
\path[clip] (  0.00,  0.00) rectangle (251.50,179.23);
\definecolor[named]{drawColor}{rgb}{0.00,0.00,0.00}

\path[draw=drawColor,line width= 0.6pt,line join=round] ( 38.93, 46.40) --
	(250.65, 46.40);
\end{scope}
\begin{scope}
\path[clip] (  0.00,  0.00) rectangle (251.50,179.23);
\definecolor[named]{drawColor}{rgb}{0.00,0.00,0.00}

\path[draw=drawColor,line width= 0.6pt,line join=round] ( 63.36, 42.13) --
	( 63.36, 46.40);

\path[draw=drawColor,line width= 0.6pt,line join=round] ( 95.93, 42.13) --
	( 95.93, 46.40);

\path[draw=drawColor,line width= 0.6pt,line join=round] (128.50, 42.13) --
	(128.50, 46.40);

\path[draw=drawColor,line width= 0.6pt,line join=round] (161.08, 42.13) --
	(161.08, 46.40);

\path[draw=drawColor,line width= 0.6pt,line join=round] (193.65, 42.13) --
	(193.65, 46.40);

\path[draw=drawColor,line width= 0.6pt,line join=round] (226.22, 42.13) --
	(226.22, 46.40);
\end{scope}
\begin{scope}
\path[clip] (  0.00,  0.00) rectangle (251.50,179.23);
\definecolor[named]{drawColor}{rgb}{0.00,0.00,0.00}

\node[text=drawColor,rotate= 30.00,anchor=base east,inner sep=0pt, outer sep=0pt, scale=  1.00] at ( 66.81, 33.32) {all-2-inv};

\node[text=drawColor,rotate= 30.00,anchor=base east,inner sep=0pt, outer sep=0pt, scale=  1.00] at ( 99.38, 33.32) {rand-10x3-inv};

\node[text=drawColor,rotate= 30.00,anchor=base east,inner sep=0pt, outer sep=0pt, scale=  1.00] at (131.95, 33.32) {all-3-inv};

\node[text=drawColor,rotate= 30.00,anchor=base east,inner sep=0pt, outer sep=0pt, scale=  1.00] at (164.52, 33.32) {rand-8x4-inv};

\node[text=drawColor,rotate= 30.00,anchor=base east,inner sep=0pt, outer sep=0pt, scale=  1.00] at (197.09, 33.32) {all-4-inv};

\node[text=drawColor,rotate= 30.00,anchor=base east,inner sep=0pt, outer sep=0pt, scale=  1.00] at (229.66, 33.32) {rand-7x5-inv};
\end{scope}
\begin{scope}
\path[clip] (  0.00,  0.00) rectangle (251.50,179.23);
\definecolor[named]{drawColor}{rgb}{0.00,0.00,0.00}

\node[text=drawColor,rotate= 90.00,anchor=base,inner sep=0pt, outer sep=0pt, scale=  1.00] at (  7.08,112.39) {performance};
\end{scope}
\end{tikzpicture}